\documentclass[format=acmsmall, review=false, screen=true]{acmart}

\usepackage{booktabs} % For formal tables

\usepackage[ruled]{algorithm2e} % For algorithms
\usepackage{graphicx}
\usepackage{subcaption}
\usepackage{color}
\usepackage{bm}
\usepackage{multirow}
\usepackage{url}
\graphicspath{{./figs/}}
\usepackage{amsthm}
\usepackage{subcaption}
\usepackage{amsmath,amssymb,amsfonts}
\usepackage{textcomp}
\usepackage{dsfont}
\usepackage{footnote}
\makesavenoteenv{tabular}
\makesavenoteenv{table}

\usepackage{lineno}
\usepackage{enumitem}
\usepackage{tikz-qtree}
\usepackage{tikz-qtree-compat}

\usepackage{color}
\usepackage{tikz}

\usepackage{ upgreek }
\usetikzlibrary{shadows,trees}

\usetikzlibrary{shapes,decorations,arrows,calc,arrows.meta,fit,positioning}
\tikzset{
	-Latex,auto,node distance =1 cm and 1 cm,semithick,
	state/.style ={ellipse, draw, minimum width = 0.7 cm},
	point/.style = {circle, draw, inner sep=0.04cm,fill,node contents={}},
	bidirected/.style={Latex-Latex,dashed},
	el/.style = {inner sep=2pt, align=left, sloped}
}

\def\independenT#1#2{\mathrel{\rlap{$#1#2$}\mkern2mu{#1#2}}}
\newcommand\independent{\protect\mathpalette{\protect\independenT}{\perp}}
\theoremstyle{definition}
\newtheorem{definition}{Definition}

\makeatletter
\newcommand{\distas}[1]{\mathbin{\overset{#1}{\kern\z@\sim}}}%

\newsavebox{\mybox}\newsavebox{\mysim}
\newcommand{\distras}[1]{%
	\savebox{\mybox}{\hbox{\kern3pt$\scriptstyle#1$\kern3pt}}%
	\savebox{\mysim}{\hbox{$\sim$}}%
	\mathbin{\overset{#1}{\kern\z@\resizebox{\wd\mybox}{\ht\mysim}{$\sim$}}}%
}
\SetAlFnt{\small}
\SetAlCapFnt{\small}
\SetAlCapNameFnt{\small}
\SetAlCapHSkip{0pt}
\IncMargin{-\parindent}

\received{ }
\received[revised]{ }
\received[accepted]{ }
\begin{document}
	% Title portion. Note the short title for running heads
	\title[]{A Survey of Learning Causality with Data: \\Problems and Methods}
	
	%\title[]{Learning about Causality from Data: \\A Survey of Problems and Methods}
	
	\author{Ruocheng Guo}
	\orcid{}
	\affiliation{%
		\institution{Computer Science and Engineering, Arizona State University}
		\city{Tempe}
		\state{AZ}
		\postcode{85281}}
	\email{rguo12@asu.edu}
	\author{Lu Cheng}
	\orcid{}
	\affiliation{%
		\institution{Computer Science and Engineering, Arizona State University}
		%\streetaddress{}
		\city{Tempe}
		\state{AZ}
		\postcode{85281}}
	\email{lcheng35@asu.edu}
	\author{Jundong Li}
	\orcid{}
	\affiliation{%
		\institution{Department of Electrical and Computer Engineering, Computer Science \& School of Data Science, University of Virginia, USA}
		%\streetaddress{}
		\city{Charlottesville}
		\state{VA}
		\postcode{85281}}
	\email{jundongl@asu.edu}
	\author{P. Richard Hahn}
	\orcid{}
	\affiliation{%
		\institution{Department of Mathematics and Statistics, Arizona State University}
		%\streetaddress{}
		\city{Tempe}
		\state{AZ}
		\postcode{85281}}
	\email{prhahn@asu.edu}
	\author{Huan Liu}
	\orcid{}
	\affiliation{%
		\institution{Computer Science and Engineering, Arizona State University}
		%\streetaddress{}
		\city{Tempe}
		\state{AZ}
		\postcode{85281}}
	\email{huan.liu@asu.edu}

	\begin{abstract}
This work considers the question of how convenient access to copious data impacts our ability to learn causal effects and relations.
In what ways is learning causality in the era of big data different from -- or the same as -- the traditional one? To answer this question, this survey provides a comprehensive and structured review of both traditional and frontier methods in learning causality and relations along with the connections between causality and machine learning.
This work points out on a case-by-case basis how big data facilitates, complicates, or motivates each approach.
\end{abstract}



% 	\keywords{}
	\setcopyright{acmcopyright}
\acmJournal{CSUR}
\acmYear{2020} \acmVolume{1} \acmNumber{1} \acmArticle{1} \acmMonth{1} \acmPrice{15.00}\acmDOI{10.1145/3397269}
	
	\maketitle
	%\tableofcontents
	\renewcommand{\shortauthors}{R. Guo et al.}

\section{Introduction}
\label{sec:intro}
Causality is a generic relationship between an effect and the cause that gives rise to it.
It is hard to define, and we often only know intuitively about causes and effects.
Because it rained, the streets were wet. Because the student did not study, he did poorly on the exam. Because the oven was hot, the cheese melted on the pizza.
When it comes to learning causality with data, we need to be aware of the differences between statistical associations and causations.
For example, when the temperatures are hot, the owner of an ice cream shop may observe high electric bills and also high sales.
Accordingly, she would observe a strong association between the electric bill and the sales figures, but the electric bill was not {\em causing} the high sales --- leaving the lights on in the shop over night would have no impact on sales. 
In this case, the outside temperature is the common cause of both the high electric bill and the high sales numbers, and we say that it is a {\em confounder} of the causality of the electricity usage on the ice cream sales. 
%
%In particular, the association between electric bill and sale numbers would change dramatically if we considered it separately for days with approximately the same outside temperature.
%; this is known as {\em adjustment for confounding}, one of the key ideas underlying causal analysis.
%In this case, temperature is the \textit{confounder} which is the cause for both electric bill and sales because we can imagine that once temperature drops, the electric bill and customer flow would also decrease.
%%
%However, the relationship between electric bill and customer flow is just a statistical association, because owners of ice cream shops cannot rely on paying more electric bills to attract more customers.
%
%Formally, causality can be defined as:
%\begin{definition}{\textbf{Causality (Causation, Causal Effect or Causal Relationship)}}
%	\textit{We say that there is a causality of a variable $A$ on another variable $B$ if and only if (iff.) the value of $B$ would change by modifying the value of $A$.}
%\end{definition}
%

The ability to learn causality is considered as a significant component of human-level intelligence and can serve as the foundation of AI~\cite{pearl2018theoretical}.
Historically, learning causality has been studied in a myriad of high-impact domains including education~\cite{lalonde1986evaluating,dehejia1999causal,heckerman2006bayesian,hill2011bayesian}, medical science~\cite{mani2000causal,cross2013identification}, economics~\cite{imbens2004nonparametric}, epidemiology~\cite{hernan2000marginal,robins2000marginal,hernan2018causal}, meteorology~\cite{ebert2012causal}, and environmental health~\cite{li2014discovering}.
%
%Historically, regression adjustment has been viewed as less-than-ideal because, in principle, no matter what you have measured, some unmeasured common cause (a so-called lurking confounder) might be the true explanation of the patterns in the data.
%Consequently, the ``gold standard'' in many fields is the randomized controlled trial (RCT);  see \cite{cook2002experimental} for a review. To take a prototypical example, to study the efficacy of a drug, a patient would be randomly assigned to take the drug or not, which would guarantee that --- {\em on average} --- the treated group and the un-treated (control) group are equivalent (balanced) in all relevant respects, ruling out the possibility of lurking confounders. Then, the impact of the drug on some health outcome --- say, the duration of a migraine headache --- can be measured by comparing the average outcome of the treatment group to that of the control group.
%
Limited by the amount of data, solid prior causal knowledge was necessary for learning causality.
Researchers performed studies on data collected through carefully designed experiments where solid prior causal knowledge is of vital importance~\cite{heckerman2006bayesian}. 
Taking the \textit{randomized controlled trials} as an example~\cite{cook2002experimental}, to study the efficacy of a drug, a patient would be randomly assigned to take the drug or not, which would guarantee that --- {\em on average} --- the treated group and the un-treated (control) groups are equivalent in all relevant respects, while ruling out the influence of any other factors.
%
%Thus, what makes a difference between the patients who took the medicine and those who took the placebo can only be the effect of medicine $A$.
Then, the impact of the drug on some health outcome --- say, the duration of a migraine --- can be measured by comparing the average outcome of the two groups.
%

%The first one is where experiments are designed to collect data by excluding anything but the causality researchers are interested in.
%
%In the second approach, with limited data, researchers rely on their solid prior knowledge about causality and then query the data for what needs to be learned.
%It can be the factor which enables human beings to build AI that can beat top-tier human players in Go~\cite{silver2016mastering} and Dota2~\cite{brockman2016} only 75 years after the first work about AI~\cite{mcculloch1943logical} was published~\cite{pearl2018theoretical}.
%The problems studied in the literature of learning causality with data falls into two categories: \emph{learning causal effect} and \textit{learning causal relations}.
%
The purpose of this survey is to consider what new possibilities and challenges arise for learning about causality in the era of big data. As an example, consider that the possibility of unmeasured confounders might be mitigated by the fact that a large number of features can be measured. %Similarly, with data from a truly large scale randomized study it becomes possible to consider the specific impact the intervention has on subgroups of the study population (this is referred to as estimating heterogeneous treatment effects). 
So, first, we aim to answer causal questions using big data. For example, do positive Yelp\footnote{https://www.yelp.com/} reviews drive customers to restaurants, or do they merely reflect popularity? 
This causal question can be addressed with data from an extensive database of Yelp.
Second, answering causal questions with big data leads to some unique new problems. For example, public databases or data collected via web crawling or application program interfaces (APIs) are unprecedentedly large, we have little intuition about what types of bias a dataset can suffer from --- the more plentiful data makes it more mysterious and, therefore, harder to model responsibly. 
At the same time, causal investigation is made more challenging by the same fundamental statistical difficulties that big data poses for other learning tasks (e.g., prediction). Perhaps the most notable example is the high-dimensionality of modern data~\cite{li2017feature}, such as text data~\cite{imai2013estimating}.

Efforts have been made to the intersection between big data and causality.
Examples include but are not limited to those discussed in \cite{mooij2016distinguishing,aral2017exercise,eggers2018regression,taddy2016scalable,guo2019learning}.
The goal of this survey is to provide a comprehensive and structured review of both traditional and frontier methods in learning causality, as well as discussion about some open problems.
We do not assume that a target audience may be familiar with learning causality.

\subsection{Overview and Organization}
Broadly, machine learning tasks are either {\em predictive} or {\em descriptive} in nature.
But beyond that we may want to understand something {\em causal}, imagining that we were able to modify some variables and rerun the data-generating process. These types of questions can also take two (related) forms: 1) How much would some specific variables (features or the label) change if we manipulate the value of another specific variable? and 2) By modifying the value of {\em which} variables could we change the value of another variable? These questions are referred to as \emph{causal inference} and \emph{causal discovery} questions, respectively~\cite{gelman2011causality,peters2017elements}.
For learning causal effects, we investigate to what extent manipulating the value of a potential cause would influence a possible effect.
Following the literature, we call the variable to be manipulated as \textit{treatment} and the variable for which we observe the response as \textit{the outcome}, respectively. One typical example is that \emph{how much do hot temperatures raise ice cream sales}.
%
%How much do hot temperatures raise ice cream sales, for example. 
%
For learning causal relations, researchers attempt to determining whether there exists a causal relationship between a variable and another. In our temperature and ice cream example, it is clear that ice cream sales do not cause high temperatures, but in other examples it may not be clear. For example, we may be interested in investigating the question like \emph{whether a genetic disposition towards cancer should be responsible for individuals taking up smoking?}
%

%Given a dataset, %the task of supervised learning is to build a predictive model from labeled training data such that it can predict the labels for unlabeled test data.
%
%beyond the conventional machine learning tasks, there are two more challenging questions we can ask by imaging as if we were able to intervene and rerun the data-generating process: how much a variable (a feature or the label) will change if we manipulate the value of another interesting variable by a certain amount? By modifying the value of which variables could we change the value of a given variable?
%
%In the literature of learning causality, these two questions correspond to the most important problems studied in the domain of \emph{learning causal effect} and \emph{learning causal relationship}~\cite{gelman2011causality,PetJanSch17}, respectively.
%
%They are traditionally referred to as \emph{causal inference} and \emph{causal discovery}, respectively.
%
%For learning causal effect (causal inference), we investigate to what extent manipulating the value of a potential cause would influence a possible effect.
%
%Following the literature, we call the variable which we wish to manipulate as \textit{treatment} and the variable for which we observe the response as \textit{the outcome}, respectively.
%
%In learning causal relations (causal discovery), researchers target at determining whether there exists a causal relationship between a variable and another, for example, distinguishing cause from effect~\cite{mooij2016distinguishing}.

In this survey, we aim to provide a comprehensive review on how to learn causality from massive data.
Below, we present an outline of the topics that are covered in this survey.
%
%Here, we first present an outline of the topics that are covered in this survey as below.
%
%\begin{enumerate}[labelindent=0pt,labelwidth=0.75em,leftmargin=!]
%\item \textbf{Preliminaries for learning causality with data} (Section~\ref{sec:model})
%\begin{enumerate}[label=(\alph*)]
%	\item Structural causal models
%	\item The potential outcome framework
%\end{enumerate}
%\item \textbf{Causal inference: learning causal effects} (Section~\ref{sec:infer})
%	\begin{enumerate}[label=(\alph*)]
%		\item Learning causal effects for data with back-door criterion
%		\item Learning causal effects for data with unobserved confounders
%	\end{enumerate}
%\item \textbf{Causal discovery: learning causal relations} (Section~\ref{sec:dis})
%\begin{enumerate}[label=(\alph*)]
%	\item Learning causal relations with i.i.d. data
%	\item Learning causal relations with time series data
%\end{enumerate}
%\item \textbf{The connection to machine learning} (Section~\ref{sec:ml})
%\begin{enumerate}[label=(\alph*)]
%	\item Supervised and semi-supervised learning
%	\item Domain adaptation
%	\item Reinforcement learning
%\end{enumerate}
%\end{enumerate}
%
First, in Section~\ref{sec:model}, we introduce the preliminaries of learning causality from data for both causal inference or causal discovery. We focus on the two formal frameworks, namely \emph{structural causal models}~\cite{Pearl2009} and the \emph{potential outcome framework}~\cite{neyman1923applications,rubin1974estimating}.
%
%In particular, we describe how causality is mathematically formulated within these two frameworks
%
%Practically, we describe a generalization of these models that can be useful for practice.
%
% Next, in Section~\ref{sec:infer} and~\ref{sec:dis} we go over the most common methodologies for learning causality from data.
%
% Specifically, in these two sections, the methods are categorized by the types of data they can handle.
%
Section~\ref{sec:infer} focuses on the methods that are developed for the problem of learning causal effects (causal inference).
Based on different types of data, these methods fall into three categories: methods for data without and with unobserved confounders, and advanced methods for big data.
In Section~\ref{sec:dis}, the widely used methods for learning causal relations are discussed.
After introducing traditional methods, we describe advanced methods addressing special challenges in big data.
% According to the data type, we first cover the methods for discovering causal relations between variables in i.i.d. data. Then, we describe the methods that can tackle the inter-dependencies in time series data.
%
Afterwards, in Section~\ref{sec:ml}, we discuss recent research that connects learning causality and machine learning.
We examine how the research in three subareas, supervised and semi-supervised learning, domain adaptation and reinforcement learning, can be connected to learning causality.

\subsection{Data for Learning Causality}
We discuss data for learning causality\footnote{The data and algorithm indexes for learning causality are covered in the Appendix}.
A comprehensive introduction of such data can be found in~\cite{cheng2019practical}.
{\color{black} Although \textit{interventional data} and a mixture of interventional and observational data are used in the literature~\cite{yin2019identification,silva2016observational,kallus2018removing,hauser2015jointly}, we focus on \textit{observational data} in this survey to take advantage of pervasive big data.
In observational data, the value of a variable is determined by its causes. In contrast, in interventional data, there exists at least one variable whose value is set through intervention.
For example, to study the causal effect of Yelp rating on customer flows of restaurants, we can either use existing records (observational data) or collect interventional data via manipulating the ratings of restaurants.}
%
% We start with the data types and methods for learning causal effects and then cover those for learning causal relations.
%

{\color{black} The ground truth used to train or evaluate causal learning methods often includes causal effects or causal relations.
Ground truth of average causal effects can be obtained via randomized experiments.
For example, the average effect of a new feature in a recommendation system is often estimated through an A/B test~\cite{yin2019identification}.
However, it is often not possible to collect ground truth of individual causal effects via randomized experiments as it requires knowledge of the \textit{counterfactual}~\cite{pearl2009causality}.
In practice, such ground truth is often acquired through simulations based on domain knowledge~\cite{johansson2016learning,louizos2017causal}.
%
%Often, ground truth of causal effects is not required for training (HL: What do you mean by this?).
%
Ground truth of causal relations is often obtained through prior knowledge (e.g., certain mutations of genes can cause diseases).}
%
% We often do not need such ground truth in training to obtain models that work for observational data.}

\noindent\textbf{Data for Learning Causal Effects.}
Here, we review the types of data, the problems that can be studied if the data is given, and the methods that can provide practical solutions.
We list three types of data for learning causal effects.
First, a standard dataset for learning causal effects $(\bm{X},\bm{t},\bm{y})$ includes feature matrix $\bm{X}$, a vector of treatments $\bm{t}$ and outcomes $\bm{y}$.
%
% This type of data is similar to what is often used for supervised learning.
%
We are particularly interested in the causal effect of one variable $t$ (treatment) on another variable $y$ (outcome).
For the second type, there is auxiliary information about inter-dependence or interference between units such as links or temporal inter-dependencies between different data units, represented by a matrix $\bm{A}$. Examples of this type of data include attributed networks~\cite{li2019adaptive,zhou2018sparc}, time series~\cite{eichler2012causal}, and marked temporal point process~\cite{guo2018initiator}.
Moreover, the third type of data often comes with unobserved confounders, we need the help of special causal variables, including the \emph{instrumental variable (IV)}, the \emph{mediator}, or the \emph{running variable}. These variables are defined by causal knowledge, thus specific methods can be applied for learning causal effect for such types of data.

\noindent\textbf{Data for Learning Causal Relations.}
We describe two types of data for learning causal relations (causal discovery).
The first type is the multivariate data $\bm{X}$ along with a ground truth causal graph $G$ for evaluation, with which we learn the causal graph. A special case is the bivariate data and the task reduces to distinguishing the cause from the effect~\cite{mooij2016distinguishing}.
The causal graph is often defined by prior knowledge and could be incomplete.
The second type of data for learning causal relations is the multivariate time series data which also comes with a ground truth causal graph. The task is to learn causal relations among the variables~\cite{gong2017causal}.
{\color{black} Although the ground-truth causal graph is often unique~\cite{peters2017elements,gong2017causal}, many methods output a set of candidate causal graphs.}  %as it is quite challenging to learn causal relations.}

\begin{table}[!tb]
	\centering
	\footnotesize
	\begin{tabular}{ |l|l|l|l| }
		\hline
		Problems & Data & Example Datasets & Methods\\
		\hline
		\multirow{8}{1.5cm}{Learning causal effects} &\multirow{4}{4cm}{Datasets contain features, treatment and outcome $(\bm{X},\bm{t},\bm{y})$.} & \multirow{4}{2.5cm}{IHDP, Twins
% 		\footnote{https://github.com/AMLab-Amsterdam/CEVAE/tree/master/datasets/IHDP}
		,\\
% 			IHDP2\footnote{https://math.la.asu.edu/$\sim$prhahn/},
			%Amazon\footnote{https://github.com/rguo12/CIKM18-LCVA},
			Jobs
% 			\footnote{http://users.nber.org/~rdehejia/data/nswdata2.html}
			}
		& \multirow{4}{3cm}{Regression adjustment,\\ Propensity score,\\ Covariate balancing,\\Machine learning based}  \\
		&&&\\
		&&&\\
		%&&&Sparse Learning Methods&\\
		%&&&Deep Learning Methods&\ref{subsec:adv_methods}\\
		%&&&Ensemble Methods&\ref{subsec:adv_methods}\\
		&&&\\ \cline{2-4}
		%
% 		&\multirow{6}{4cm}{Non-i.i.d. data satisfies the back-door criterion: covariates, treatment, outcome and inter-dependency $(\bm{X},\bm{d},\bm{y},\bm{A})$, similar to attributed network, time series data or marked temporal point process.} & \multirow{6}{2cm}{
% 			Amazon\footnote{https://github.com/rguo12/CIKM18-LCVA}}
% 			%Job Training\footnote{text}}
% 		 & \multirow{6}{3cm}{Explicit modeling interference; Disentangle units with i.i.d representation} & \multirow{6}{1cm}{\ref{subsec:CI_non-i.i.d}}  \\
% 		&&&&\\
% 				&&&&\\
% 		&&&&\\
% 		&&&&\\
% 		&&&&\\
% 		&&&&\\ \cline{2-5}
		%
		& \multirow{4}{4.5cm}{Datasets with features, treatment, outcome and special variable(s): $(\bm{X},\bm{t},\bm{y},\bm{z})$ } & \multirow{4}{2.5cm}{1980 Census Extract,\\
			CPS Extract
% 			\footnote{https://economics.mit.edu/faculty/angrist/data1/data/angkru95}
			}  & \multirow{4}{3cm}{IV methods,\\ Front-door criterion,\\ RDD,\\Machine learning based}\\
		&&&\\
		&&&\\
				&&&\\\hline
		%
%		\multirow{5}{4.5cm}{Datasets for Causal Inference with Mediators. Each unit is a tuple of the Mediator(s), other covariates, treatment and outcome $(M,X,D,Y)$.} & \multirow{5}{1.5cm}{} & \multirow{5}{1.5cm}{Estimating ATE} & \multirow{5}{2.5cm}{Front-door Criterion} & \multirow{5}{*}{\ref{subsec:FDC}} \\
%		&&&&\\
%		&&&&\\
%		&&&&\\
%		&&&&\\ \hline
		%
%		\multirow{5}{4.5cm}{Datasets for Causal Inference with a Running Variable. Each unit is a tuple of the Running Variable and outcome $(R,Y)$.} & \multirow{5}{1.5cm}{Population-threshold RDD dataset\footnote{https://dataverse.harvard.edu/dataset.xhtml?persistentId=doi:10.7910/DVN/PGXO5O}} & \multirow{5}{1.5cm}{Estimating ATE} & \multirow{5}{2.5cm}{Sharp RDD and Fuzzy RDD.} & \multirow{5}{*}{\ref{subsec:RDD}} \\
%		&&&&\\
%		&&&&\\
%		&&&&\\
%		&&&&\\ \hline
		%
		%
		\multirow{8}{1.5cm}{Learning causal relations.}& \multirow{5}{4.5cm}{Multivariate data with causal relations, denoted by $\bm{X}$ with a causal graph $G$, including bivariate data with causal direction.} & \multirow{5}{2.5cm}{Abscisic Acid Signaling Network
% 		\footnote{https://archive.ics.uci.edu/ml/datasets/Abscisic+Acid+Signaling+Network}
, Weblogs
% \footnote{http://www.causality.inf.ethz.ch/repository.php?id=13},
\\ SIDO
% ~\footnote{http://www.causality.inf.ethz.ch/data/SIDO.html}}
}
& \multirow{8}{3cm}{Constraint-based,\\ Score-based methods,\\Algorithms for FCMs. } \\
% 		& \multirow{9}{*}{\ref{subsec:CL_IND}} \\
		&&&\\
		&&&\\
		&&&\\
		&&&\\ \cline{2-3}
		&\multirow{3}{4.5cm}{Multivariate time series $\{[x_{,1}(l),...,x_{,J}(l)]\}_{l=1}^L$ with a causal graph $G$} & \multirow{3}{1.5cm}{PROMO}
% 		\footnote{http://clopinet.com/causality/data/promo/}} 
&  \\
		&&&\\
		&&&\\
		&&&\\ \hline
%		%
%		\multirow{5}{4.5cm}{Bivariate data for Causal Learning. Each sample is a tuple of the two variables $(X_1,X_2)$. Or two univariate time series.} & \multirow{5}{1.5cm}{Database with cause-effect pairs\footnote{http://webdav.tuebingen.mpg.de/cause-effect/}} & \multirow{5}{1.5cm}{Distinguish cause from effect.} & {Additive Noise Models.} & \multirow{5}{*}{\ref{subsec:CL_IND}} \\
%		&&&Information Geometric Causal Inference&\\
%		&&&Supervised Learning Methods&\\ \hline
		%
%		\multirow{6}{4.5cm}{Multivariate data for Cause Detection. Each sample is a tuple of variables and a outcome $(X_1,...,X_N,Y)$. Only a subset of variables are the causes.} & \multirow{6}{1.5cm}{} & \multirow{6}{1.5cm}{Detection of causes among a set of variables.} & \multirow{6}{2.5cm}{} & \multirow{5}{*}{\ref{subsec:CL_SCMs}} \\
%		&&&&\\
%		&&&&\\
%		&&&&\\
%		&&&&\\
%		&&&&\\ \hline
	\end{tabular}
	\caption{Overview of this work in terms of the problems, data and methods.}
	\vspace{-0.3in}
	\label{tab:summary}
\end{table}

\subsection{Previous Work and Contributions}
There are a number of other comprehensive surveys in the area of causal learning.
Pearl~\cite{Pearl2009} aims to convey the fundamental theory of causality based on the structural causal models.
Gelman~\cite{gelman2011causality} provides high-level opinions about the existing formal frameworks and problems for causal learning.
Mooji et al.~\cite{mooij2016distinguishing} focus on learning causal relations for bivariate data.
Spirtes and Zhang~\cite{spirtes2016causal} summarize methods for learning causal relations on both i.i.d. and time series data with a focus on several semi-parametric score based methods.
Athey and Imbens~\cite{athey2015machine} describe decision trees and ensemble machine learning models for learning causal effects.
%
%Hernan and Robins~\cite{robins2000marginal} addressed the real-world problem of time-varying treatment.

Different from previous work, this survey is structured around various data types, and what sorts of causal questions can be addressed with them.
%
%We aim to provide machine learning and data mining researchers with a convenient guide on causal learning.
%
Specifically, we describe what types of data can be used for the study of causality, what are the problems that can be solved for each type of data and how they can be solved.
In doing so, we aim to provide a bridge between the areas of machine learning, data mining, and causal learning in terms of terminologies, data, problems and methods.

\subsection{Running Example}
We consider a study of how Yelp ratings influence potential restaurant customers~\cite{Anderson2012}.
Yelp is a website where customers can share their reviews of a certain goods and services.
%, car rental service and hair salon.
%
Each review includes an integer rating from 1 to 5 stars.
For our purposes, the Yelp rating is our \textit{treatment} variable and the number of customers (in some well-defined period) is the \textit{outcome} variable.
For simplicity, we assume that these variables are binary. A restaurant receives the treatment $t=1$ if its rating is above some threshold; otherwise, it is under control treatment $t=0$.
For the outcome, $y=1$ means a restaurant is completely booked and $y=0$ means it is not. 

\section{Preliminaries}
\label{sec:model}

Here, we present the preliminaries for two fundamental frameworks: structural causal models and the potential outcome framework.
To formulate causal knowledge, we need \textit{causal models}.
A causal model is a mathematical abstraction that quantitatively describes the causal relations between variables.
First, causal assumptions or prior causal knowledge can be represented by an incomplete causal model.
Then, what is missing can be learned from data.
The two most well-known causal models are the structural causal models (SCMs)~\cite{pearl2009causality} and the potential outcome framework~\cite{neyman1923applications,rubin1974estimating}.
They are considered as the foundations because they enable a consistent representation of prior causal knowledge, assumptions, and estimates.

We present the terminologies and notations that are used throughout this survey.
Table~\ref{tab:terms} displays key nomenclature.
In this survey, a lowercase letter, say $x$, denotes a specific value of a corresponding random variable (or RV), $X$.
Bold lowercase letters denote vectors or sets (e.g., $\bm{x}$) and bold uppercase letters signify matrices (e.g., $\bm{X}$).
Calligraphic uppercase letters such can signify special sets such as sets of nodes $\mathcal{V}$ and edges $\mathcal{E}$ in a graph $G$. 
$\bm{X}$ and $\bm{x}_i$ present features for all instances and that for the $i$-th instance, respectively.
Without specification, the subscripts denote the instance and the dimension.
For example, $\bm{x}_i$ denotes features of the $i$-th instance and $x_{\ast,j}$ signifies the $j$-th feature.
The letter $t$ will be used to denote the treatment variable; in this work, it is often assumed to be binary and univariate.
The letter $y$ denotes the outcome variable. We use the subscript and superscript of $y$ to signify the instance and the treatment it corresponds to. When the treatment is binary, $y^1_i$ denotes the outcome when the instance $i$ is treated ($t_i=1$).
The letter $\tau$ denotes various treatment effects, {\color{black} defined as a change in the outcome variable for different levels of treatment}.

\begin{table}[tb]
	\centering
\tiny
	\caption{Nomenclature}
	\begin{tabular}{ |l|l|p{5.1cm}| }
		\hline
		\multicolumn{3}{ |c| }{Nomenclature} \\
		\hline
		Terminology & Alternatives & Explanation \\ \hline
		causality & causal relation, causation & \multirow{1}{*}{causal relation between variables} \\ \hline
		causal effect & & the strength of a causal relation \\ \hline
		instance & unit, sample, example & an independent unit of the population\\ \hline
		\multirow{2}{*}{features} & covariates, observables & \multirow{2}{*}{variables describing instances}\\
		& pre-treatment variables& \\\hline
		\multirow{2}{*}{learning causal effects} & forward causal inference & \multirow{2}{5cm}{identification and estimation of causal effects} \\
		& forward causal reasoning & \\ \hline
		\multirow{3}{*}{learning causal relations} & causal discovery& \multirow{3}{5cm}{inferring causal graphs from data} \\
		& causal learning &  \\
		& causal search &  \\ \hline
		causal graph & causal diagram & a graph with variables as nodes and causality as edges\\ \hline
		confounder & confounding variable & a variable causally influences both treatment and outcome\\ \hline
	\end{tabular}
	\label{tab:terms}
\end{table}

\subsection{Structural Causal Models}
\begin{figure}[tb]
	\begin{subfigure}[b]{0.32\textwidth}
		\centering
		\begin{tikzpicture}
		
		\node[state] (x) at (0,0) {$x$};
		
		\node[state] (z) [right =of x] {$z$};
		\node[state] (y) [right =of z] {$y$};
		
		% Directed edge
		\path (x) edge (z);
		\path (z) edge (y);
		
		% Bidirected edge
		
		\end{tikzpicture}
		\caption{Chain}
		\label{fig:DAG1}
	\end{subfigure}
	~ %add desired spacing between images, e. g. ~, \quad, \qquad, \hfill etc.
	\begin{subfigure}[b]{0.32\textwidth}
		\centering
		\begin{tikzpicture}
		
		\node[state] (x) at (0,0) {$x$};

		\node[state] (z) [right =of x] {$z$};
		\node[state] (y) [right =of z] {$y$};
		
		% Directed edge
		\path (z) edge (x);
		\path (z) edge (y);

		\end{tikzpicture}
				\caption{Fork}
		\label{fig:DAG2}
	\end{subfigure}
	\begin{subfigure}[b]{0.32\textwidth}
		\centering
		\begin{tikzpicture}
% x node set with absolute coordinates
\node[state] (x) at (0,0) {$x$};

\node[state] (z) [right =of x] {$z$};
\node[state] (y) [right =of z] {$y$};

% Directed edge
\path (x) edge (z);
\path (y) edge (z);

\end{tikzpicture}
				\caption{Collider}
		\label{fig:DAG3}
	\end{subfigure}
\caption{Three typical DAGs for conditional independence}
\vspace{-0.2in}
    \label{fig:DAGs}
\end{figure}
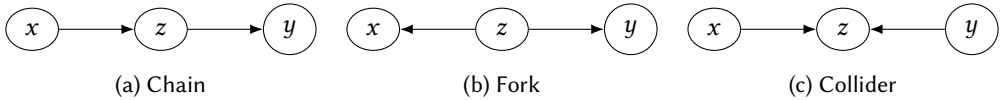
Structural causal models (SCMs) provide a comprehensive theory of causality~\cite{Pearl2009}.
An SCM often consists of two components: the \textit{causal graph} (causal diagram) and the \textit{structural equations}.

\noindent\textbf{Causal Graphs.} A causal graph forms a special class of Bayesian network with edges representing the causal effect, thus it inherits the well defined conditional independence criteria.
\begin{definition}{\textbf{Causal Graph.}}
	\textit{A causal graph $G=(\mathcal{V},\mathcal{E})$ is a directed graph that describes the causal effects between variables, where $\mathcal{V}$ is the node set and $\mathcal{E}$ the edge set. In a causal graph, each node represents a random variable including the treatment, the outcome, other observed and unobserved variables. A directed edge $x\rightarrow y$ denotes a causal effect of $x$ on $y$.}
\end{definition}

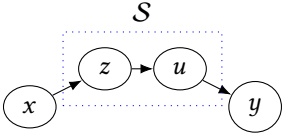
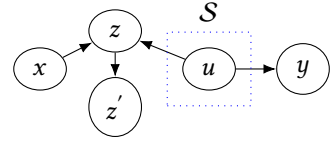
\begin{figure}[tb]
	\begin{subfigure}[tb]{0.4\textwidth}
		\centering
		\begin{tikzpicture}
		% x node set with absolute coordinates
		\node[state] (x) at (0,0) {$x$};
		
		\node[state] (z) at (1,0.5) {$z$};
		\node[state] (u) at (2,0.5) {$u$};
		\node[state] (y) at (3,0) {$y$};
		
		% Directed edge
		\path (x) edge (z);
		\path (z) edge (u);
		\path (u) edge (y);
		
		\node[draw=blue,dotted,fit=(z) (u), inner sep=0.2cm,label=$\mathcal{S}$] (machine) {};
		
		% Bidirected edge
		\end{tikzpicture}
		\caption{Conditioning on $\mathcal{S}$ blocks the node $z$ as $z\in\mathcal{S}$ and $z$ is not a collider.}
		\label{fig:block1}
	\end{subfigure}
\hfill
\begin{subfigure}[tb]{0.4\textwidth}
	\centering
	\begin{tikzpicture}
	\node[state] (x) at (0,0) {$x$};
	
	\node[state] (z) at (1,0.5) {$z$};
	\node[state] (z1) at (1,-0.5) {$z^{'}$};
	\node[state] (u) at (2.25,0) {$u$};
	\node[state] (y) at (3.5,0) {$y$};
	
	% Directed edge
	\path (x) edge (z);
	\path (u) edge (z);
	\path (z) edge (z1);
	\path (u) edge (y);
	
	\node[draw=blue,dotted,fit=(u), inner sep=0.2cm,label=$\mathcal{S}$] (machine) {};
	
	\end{tikzpicture}
	\caption{Conditioning on $\mathcal{S}$ blocks $z$ as $z$ is a collider and neither $z$ nor $z^{'}$ is in $\mathcal{S}$.}
	\label{fig:block2}
\end{subfigure}
\caption{Examples of $z$ being blocked by conditioning on $\mathcal{S}$}
\end{figure}

A \textit{path} is a sequence of directed edges and a \textit{directed path} is a path whose edges point to the same direction.
In this work, as is common in the field, we only consider \textit{directed acyclic graphs} (DAGs) where no directed path starts and terminates at the same node.
Given a SCM, the conditional independence embedded in its causal graph provides sufficient information confirm whether it satisfies the criteria such that we can apply certain causal inference methods.
To understand the conditional independence, here, we briefly review the concept of \textit{dependency-separation} (d-separation) based on the definition of \textit{blocked} path.
Fig.~\ref{fig:DAGs} shows three typical DAGs.
In the \textit{chain} (Fig.~\ref{fig:DAG1}), $x$ causally affects $y$ through its influence on $z$.
In the \textit{fork} (Fig.~\ref{fig:DAG2}), $z$ is the common cause of both $x$ and $y$.
In this case, $x$ is associated with $y$ but there is no causation between them.
When $z$ is a \textit{collider} node (see Fig.~\ref{fig:DAG3}), both $x$ and $y$ cause $z$ but there is no causal effect or association between $x$ and $y$.
In the chain and fork, the path between $x$ and $y$ is blocked if we condition on $z$, which can be denoted as $x \independent y | z$.
Contrarily, in a collider (Fig.~\ref{fig:DAG3}), conditioning on $z$ introduces an association between $x$ and $y$, i.e., $x \independent y$ and $x \not \independent y|z$.
Generally, we say conditioning on a set of nodes $\mathcal{Z}$ blocks a path $p$ iff there exists at least one node $z\in\mathcal{Z}$ in the path $p$.
\begin{definition}\textbf{Blocked.}
	\textit{We say a node $z$ is blocked by conditioning on a set of nodes $\mathcal{S}$ if one of the two conditions is satisfied: (1) $z \in \mathcal{S}$ and $z$ is not a collider node (Fig.~\ref{fig:block1}); (2) $z$ is a collider node, $z\not\in\mathcal{S}$ and no descendant of $z$ is in $\mathcal{S}$ (Fig.~\ref{fig:block2}).}
\end{definition}

{ \color{black} With this definition, we say a set of variables $\mathcal{S}$ $d$-separates two variables $x$ and $y$ iff $\mathcal{S}$ blocks all paths between them.
$d$-separation plays a crucial role in explaining causal concepts.
The \textit{Causal Markovian condition} is often assumed in SCMs, which means we can factorize the joint distribution represented by a \textit{Markovian} SCM of variables $\mathcal{V} = \left\{x_{\ast,1},...,x_{\ast,J}\right\}$ with:
\begin{equation}
P(x_{\ast,1},...,x_{\ast,J}) = \prod_{j} P(x_{\ast,j}|\bm{Pa}_{\ast,j},\epsilon_{\ast,j}),
\end{equation}
where $\bm{Pa}_{\ast,j}$ denotes the set of parent variables of $x_{\ast,j}$, each of which has an arrow in $x_{\ast,j}$.
Moreover, $\epsilon_{\ast,j}$ is the noise which represents the causal effect of unobserved variables on $x_{\ast,j}$.
Here, we introduce the key concepts of learning causality through a toy SCM which embeds causal knowledge for the Yelp example~\cite{Anderson2012}.
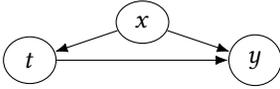
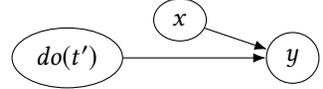
\begin{figure}
\begin{subfigure}[tb]{0.4\textwidth}
	\centering
	\begin{tikzpicture}
	% x node set with absolute coordinates
	\node[state] (d) at (0,0) {$t$};
	
	\node[state] (x) at (1.5,0.5) {$x$};
	%\node[state,rectangle] (s) at (-1.5,1) {$S$};
	%\node[state] (u) at (2,0.5) {$U$};
	\node[state] (y) at (3,0) {$y$};
	
	% Directed edge
	\path (x) edge (d);
	\path (x) edge (y);
	\path (d) edge (y);
	
	\end{tikzpicture}
	\caption{An SCM without intervention.}
	\label{fig:SCM1}
\end{subfigure}
\hfill
\begin{subfigure}[tb]{0.4\textwidth}
		\centering
	\begin{tikzpicture}
\node[state] (d) at (0,0) {$do(t')$};

\node[state] (x) at (1.5,0.5) {$x$};
\node[state] (y) at (3,0) {$y$};

% Directed edge

\path (x) edge (y);
\path (d) edge (y);

% Bidirected edge
\end{tikzpicture}
\caption{An SCM under the intervention $do(t)$.}
\label{fig:SCM2}
\end{subfigure}
\caption{SCMs without and under intervention $do(t')$ for the Yelp example, where $x$, $t$ and $y$ denote restaurant category, Yelp rating and customer flow.}
% \vspace{-7pt}
\end{figure}
In Fig.~\ref{fig:SCM1}, there are three random variables, i.e., the restaurant category $x$ (\textit{confounder}), Yelp star rating $t$ (treatment) and customer flow $y$ (outcome).
The three directed edges represent the three causal effects:
% \begin{enumerate}
	(1) Restaurant category influences its Yelp rating. For example, the average rating of fast food restaurants is lower than that of high-end restaurants.
	(2) Restaurant category also influences its customer flow. For example, the average customer flow of fast food restaurant is higher than that of high-end restaurants.
	(3) Yelp rating of a restaurant influences its customer flow.
% \end{enumerate}}
%
}

\noindent\textbf{Structural Equations.} Given a causal graph along with a set of structural equations, we can specify the causal effects signified by the directed edges.
A set of non-parametric structural equations can quantify the three causal effects shown in the causal graph in Fig.~\ref{fig:SCM1} as:
\begin{equation}
x = f_x(\epsilon_x), \; t = f_t(x,\epsilon_t), \; y = f_y(x,t,\epsilon_y).
\label{eq:SE}
\end{equation}
In Eq.~\ref{eq:SE}, $\epsilon_x$, $\epsilon_t$ and $\epsilon_y$ denote the ``noise'' of the observed variables, concieved as \textit{exogenous} or mutually independent sources of unmeasured variation.
The noise terms represent the causal effect of unobserved variables on the variable on the LHS.
Note that for each equation, we assume that the variables on the RHS influences those on the LHS, not the other way around.
Rewriting this equation in a different order as $x = f_t^{-1}(t,\epsilon_t)$ can be misleading as it implies that Yelp rating causally influences restaurant type.
The structural equations (Eq.~\ref{eq:SE}) provide a quantitative way to represent \textit{intervention} on a variable of the corresponding causal graph (Fig.~\ref{fig:SCM1}).
The \textit{do-calculus}~\cite{pearl2009causality} of Pearl was proposed to define intervention in SCMs.
Specifically, the do-calculus introduces a new operator $do(t')$, which denotes the intervention of setting the value of the variable $t$ to $t'$.
The notation of $do(t)$ leads to a formal expression of the interventional distributions:
\begin{definition}{\textbf{Interventional Distribution (Post-intervention Distribution).}}
	\textit{The interventional distribution $P(y|do(x'))$ denotes the distribution of the variable $y$ when we rerun the modified data-generation process where the value of variable $x$ is set to $x'$.}
\end{definition}
For example, for the causal graph in Fig.~\ref{fig:SCM1}, the post-intervention distribution $P(y|do(t))$ refers to the distribution of customer flow $y$ as if the rating $t$ is set to $t'$ by intervention, where all the arrows into $t$ are removed, as shown in Fig.~\ref{fig:SCM2}.
The structural equations associated with Fig.~\ref{fig:SCM2} under the intervention on the treatment variable, denoted by $do(t')$, can be written as:
\begin{equation}
x = f_x(\epsilon_x), \;
t = t', \;
y = f_y(x,t,\epsilon_y),
\label{eq:SE1}
\end{equation}
which formulates the interventional distribution as $P(y|do(t'))=f_y(x,t,\epsilon_y)$.
Then, when it comes to the causal effect of $t$ on $y$, in the language of SCMs, the problem of calculating causal effects can be translated into queries about the interventional distribution $P(y|do(t))$ with different $t$.
Implicitly, we assume that the variables follow the same causal relations of a SCM for each instance.
Hence, SCMs enable us to define \textit{average treatment effect} (ATE).
For the running example, the ATE of Yelp rating might be defined as:
\begin{equation}
\tau(t,c) = \mathds{E}[y|do(t)]-\mathds{E}[y|do(c)], t>c,
\end{equation}
where $t$ and $c$ refer to the ratings that are considered as positive and negative, respectively.
In many cases, the treatment variable is binary, thus the ATE reduces to a value $\mathds{E}[y|do(1)]-\mathds{E}[y|do(0)]$.
{ \color{black} It is crucial to note that $P(y|do(t))$ and $P(y|t)$ are not the same, which makes calculating ATEs impossible.
This gap can give rise to \textit{confounding bias}, which results if one estimates treatment effects using $P(y | t)$ where $P(y | do(t))$ is in fact required.
The existence of a \textit{back-door path} is a common source of confounding, rendering $P(y|do(t))$ and $P(y|t)$) distinct.
An example of is the path $t\leftarrow x \rightarrow y$ in Fig.~\ref{fig:SCM1}. Observe also that randomized treatment assignment directly avoids back-door paths, side-stepping confounding bias.
We present the formal definitions of confounding bias, back-door path and confounder with do-calculus and SCMs.
\begin{definition}{\textbf{Confounding Bias.}}
	\textit{Given variables $x$, $y$, confounding bias exists for causal effect $x\rightarrow y$ iff the probabilistic distribution representing the statistical association is not always equivalent to the interventional distribution, i.e., $P(y|x)\not = P(y|do(x))$.
}
\end{definition}
\begin{definition}{\textbf{Back-door Path.}}
	\textit{Given a pair of treatment and outcome variables $(t,y)$, we say a path connecting $t$ and $y$ is a back-door path for $(t,y)$ iff it satisfies that (1) it is not a directed path; and (2) it is not blocked (it has no collider).}
\end{definition}
\begin{definition}{\textbf{Confounder (Confounding Variable).}}
	\textit{Given a pair of treatment and outcome variables $(t,y)$, we say a variable $z \not \in \left\{t,y\right\}$ is a confounder iff it is the central node of a fork and it is on a back-door path of $(t,y)$.}
\end{definition}}
In particular, in the running example, the probability distribution $P(y|t)$ results from a mixture of the causal effect $P(y|do(t))$ and the statistical associations produced by the back-door path $t\leftarrow x \rightarrow y$, where $x$ is the confounder.
To obtain unbiased estimate of causal effects from observational data requires eliminating confounding bias, a procedure referred to as \textit{causal identification}.
\begin{definition}{\textbf{Causal Identification.}}
	\textit{We say a causal effect is identified iff the hypothetical distribution (e.g., interventional distribution) that defines the causal effect is formulated as a function of probability distributions over observables.}
\end{definition}
A common way to identify causal effects in SCMs is to block the back-door paths that reflect other irrelevant causal effects.
A way to eliminate confounding bias is to estimate the causal effect within subpopulations where the instances are homogeneous w.r.t. confounding variables~\cite{Pearl2009}.
This corresponds to \textit{adjustment} on variables that satisfy the \textit{back-door criterion} for causal identification~\cite{pearl2009causality}.
Now we present a formal definition of the back-door criterion.
\begin{definition}{\textbf{Back-door Criterion.}}
	\textit{Given a treatment-outcome pair $(t,y)$, a set of features $\bm{x}$ satisfies the back-door criterion of $(t,y)$ iff conditioning on $\bm{x}$ can block all back-door paths of $(t,y)$.}
\end{definition}
A set of variables that satisfies the back-door criterion is referred to as an \textit{admissible set} or a \textit{sufficient set}.
For the running example, we are interested in the causal effect of Yelp star rating on the customer flow ($t\rightarrow y$) or equivalently the interventional distribution $P(y|do(t))$. So for causal identification, we aim to figure out a set of features that satisfies the back-door criterion for the treatment-outcome pair $(t,y)$.
For example, if restaurant category $x_{,j}$ is the only confounder for the causal effect of Yelp rating on customer flow, then $\mathcal{S}=\left\{x_{,j}\right\}$ satisfies the back-door criterion.
There are two types of data w.r.t. the back-door criterion for causal inference.
The first data type assumes that the whole set or a subsets of the features $\mathcal{S}$ satisfies the back-door criterion such that by making adjustment on $\mathcal{S}$, $P(y|do(t))$ can be identified. We will introduce methods for learning causal effects with data of this type in Section~\ref{subsec:SD}.
In the second data type, other criteria are used to identify causal effects without the back-door criterion satisfied.

\noindent\textbf{Confounding bias without back-door path.} Confounding bias may exist without back-door paths. An example is a type of selection bias~\cite{bareinboim2015recovering}, when the causal graph is $t\rightarrow z \leftarrow x \rightarrow y$ and the dataset is collected only for instances with $z_i=1$, then within this dataset, the estimated statistical association $P(y|t)$ can be non-zero although we know that there is no causal effect $t\rightarrow y$.
Selection bias can also result from adjustment on certain variables (e.g., colliders or descendants of the outcome variable).
Without knowing the complete graph, Entner et al.~\cite{entner2013data} provide a set of rules that are sufficient to decide whether a set of variables satisfy the back door criterion, or that $t$ actually has no effect on $y$.
This implies that  there is a middle ground between hoping to adjust for all and only the right stuff, and trying to learn the entire causal graph.

\noindent\textbf{Beyond do-calculus.} Do-calculus has some limitations, which mainly come from the i.i.d. assumption~\cite{Richardson}.
This implies that it is difficult to formulate individual-level hypothetical distributions with do-calculus.
Let us consider the running example, even if we could hack Yelp and let it show median rating instead of average rating, we still cannot answer questions such as what would the customer flow for a restaurant be if we had increased its rating by 0.5 star without changing the ratings of others?
In~\cite{pearl2018theoretical}, Pearl refers to the hypothetical distributions which cannot be identified through interventions as \textit{counterfactuals}.
Do-calculus, which formally represents hypothetical interventions, cannot formulate counterfactuals in SCMs.
Therefore, besides do-calculus, Pearl~\cite{Pearl2009} introduced a new set of notations. For example, $P(y^t|y',t')$ denotes the probability of the outcome  $y$ would be if the observed treatment value is $t$, given the fact that we observe $y',t'$ in the data.
In the running example, for a restaurant with rating $t'$ and customer flow $y'$, the counterfactual probability $P(y^t|y',t')$ is the distribution of the customer flow if we had observed its rating as $t$.

\subsection{Potential Outcome Framework}
\label{subsec:POF}
The potential outcome framework~\cite{neyman1923applications,rubin1974estimating} is widely used by practitioners to learn causal effects as it is defined w.r.t. a given treatment-outcome pair $(t,y)$.
A \textit{potential outcome} is defined as:
\begin{definition}
\noindent{\textbf{Potential Outcome.}}
\textit{Given the treatment and outcome $t,y$, the potential outcome of instance $i$, $y^{t}_i$, is the outcome that would have been observed if the instance $i$ had received treatment $t$.
}
\end{definition}
\noindent This framework allows a straightforward articulation of the basic challenge of causal inference~\cite{holland1986statistics}: only one potential outcome can be observed for each instance.
\noindent Using potential outcomes it is possible to define the \textit{individual treatment effect} (ITE) as the difference between potential outcomes of a certain instance under two different treatments.
ITE can be extended to ATE on arbitrary populations.
Practitioners often assume binary treatment ($t\in\left\{0,1\right\}$), where $t=1$ ($t=0$) mean that an instance is under treatment (control).
The formal definition of ITE is:
\begin{definition}{\textbf{Individual Treatment Effect.}}
	\textit{Assuming binary treatment, given an instance $i$ and its potential outcomes $y_i^t$, the individual treatment effect is defined as $\tau_{i} = y_i^{1}-y_{i}^0$.}
\end{definition}
Based on ITE, the ATE of the target population and other subpopulation average treatment effects such as \textit{conditional average treatment effect} (CATE) can be defined.
Earlier in this section, we have already defined ATE with do-calculus, here we show that ATE can also be formulated in the potential outcome framework.
Given ITEs, ATE can be formulated as the expectation of ITEs over the whole population $i=1,...,n$ as:
\begin{equation}
\label{eq:ATE}
\tau = \mathds{E}_{i}[\tau_i] = \mathds{E}_{i}[y_{i}^1-y_{i}^0] = \frac{1}{n}\sum_{i=1}^n(y_{i}^1-y_{i}^0),
\end{equation}
The ATE on subpopulations is often of interest.
An example is the \emph{conditional average treatment effect} (CATE) of instances with the same features, i.e., $\tau(\bm{x}) = \mathds{E}_{i:\bm{x}_i=\bm{x}}[\tau_i]$.

\noindent\textbf{Using the potential outcome framework to estimate treatment effects.} 
Similar to how $P(y|do(t))$ and $P(y|t)$ are distinct in the SCM framework, $P(y^t)$ and $P(y | t = 1)$ are not the same within the potential outcomes framework. First, the most fundamental assumption that is commonly used to facilitate estimation is \textit{the stable unit treatment value assumption} (SUTVA), which can be broken down into two conditions: \textit{well-defined treatment levels} and \textit{no interference}.
The condition of well-defined treatment indicates that given two different instances $i\not=j$, if the values of their treatment variable are equivalent, then they receive the same treatment.
The condition of no interference signifies that the potential outcomes of an instance is independent of what treatments the other units receive, which can be formally expressed as $y_{i}^{\bm{t}} = y_i^{t_i}$,
where $\bm{t}\in \left\{0,1\right\}^{n}$ denotes the vector of treatments for all instances.
{ \color{black} Although the condition of no interference is often assumed, there are cases when the inter-dependence between instances matters~\cite{rakesh2018linked,Toulis2018}.}
The second assumption, {\em consistency}, means that the observed outcome is independent of the how the treatment is assigned.
Finally, a commonly invoked condition is the unconfoundedness (a.k.a. ignorability), which posits that the set of confounding variables $\mathcal{S}$ can be measured.
Starting from this point, we use $\bm{s}$ to denote the vector of confounding variables.
{ \color{black} \textit{Unconfoundedness} means that the values of the potential outcomes are independent of the observed treatment, given the set of confounding variables.} Unconfoundedness is defined as:
$\label{eq:unconf}
y_i^1,y_i^0 \independent t_i | \bm{s}$,
where $\bm{s}$ denotes a vector of confounders, each element of which is a feature that causally influences both the treatment $t_i$ and the outcome $y_i^t$.
We can see that this is also an assumption defined at the individual level.
Unconfoundedness directly leads to causal identification as Pearl~\cite{Pearl2009} showed that, given Eq.~\ref{eq:unconf}, $\mathcal{S}$ always satisfies the back-door criterion of $(t,y)$. That is, under unconfoundedness we have $P(y^1 | t, \bm{s}) = P(y | t, \bm{s})$. 
A further condition $P(t=1|\bm{x})\in (0,1)$ if $P(\bm{x})>0$ is often invoked, which, when combined with ignorability, is referred to as \textit{strong ignorability}.

\noindent\textbf{Comparing SCMs and potential outcomes.} 
The two formal frameworks are logically equivalent, which means an assumption in one can always be translated to its counterpart in the other~\cite{Pearl2009}.
There are also some differences between them.
In the potential outcome framework, the causal effects of the variables other than the treatment and the special variables such as instrumental variable (see Section 3.2.1) are not defined.
This is a strength of this framework as we can model the interesting causal effects without knowing the complete causal graph~\cite{aliprantis2015distinction}.
While in SCMs, we are able to study the causal effect of any variable.
Therefore, SCMs are often preferred when learning causal relations among a set of variables~\cite{aliprantis2015distinction}. Conversely, if the goal is narrowly to estimate a given treatment effect, developing estimators can be more straightforward using the potential outcomes framework. The reader may draw their own conclusions after consulting Section 3.
\section{Learning Causal Effects}
\label{sec:infer}
In this section, we introduce methods for learning causal effects.
We aim to understand how to quantify causal effects in a data-driven way.
{\color{black} We first introduce the problem statement and evaluation metrics. Next, we review three categories of methods: those with and without unconfoundedness and advanced methods for big data.}
We define the problem of learning causal effects.
\begin{definition}{\textbf{Learning Causal Effects}}
	\textit{Given $n$ instances, $[(\bm{x}_{1},t_{1},y_1),...,(\bm{x}_{n},t_{n},y_{n})]$, learning causal effects quantifies how the outcome $y$ is expected to change if we modify the treatment from $c$ to $t$, which can be defined as $\mathds{E}[y|t]-\mathds{E}[y|c]$, where $t$ and $c$ denote a treatment and the control.}
\end{definition}
Depending on the application, we care about the causal effect for different populations.
It can be the whole population, a known subpopulation that is defined by some conditions, an unknown subpopulation or an individual.
Among all types of treatment effects, the ATE is often interesting when it comes to making decision on whether a treatment should be introduced to a population.
Furthermore, in SCMs and do-calculus, the identification of ATE only requires to query interventional distributions but not counterfactuals.
This means that ATE is often easier to identify and estimate than other types of treatment effects.
In terms of evaluation, regression error metrics such as mean absolute error (MAE) can be used to evaluate models for learning ATE.
Given the ground truth $\tau$ and the inferred ATE $\hat{\tau}$, the MAE on ATE is:
\begin{equation}
	\epsilon_{MAE\_ATE} = |\tau-\hat{\tau}|.
\end{equation}

However, when the population consists of heterogeneous groups, ATE can be misleading.
For example, Yelp rating may matter much more for restaurants in big cities than those in small towns.
Therefore, ATE can be spurious as an average of heterogeneous causal effects.
In contrast, the average should be taken within each homogeneous group.
In many cases, without knowledge about the affiliation of groups, an assumption we can make is that each subpopulation is defined by different feature values.
Thus, we can learn a function to map the features that define a subpopulation to its estimated ATE.
With this assumption, given a certain value of features $\bm{x}$ and binary treatment $t$, the CATE is a function of $\bm{x}$ and is defined as:
		\begin{equation}
		\tau(\bm{x}) = \mathds{E}[y|\bm{x},t=1] - \mathds{E}[y|\bm{x},t=0].
		\label{eq:CATE}
		\end{equation}
In this case, we assume that only the features and the treatment are two factors that determine the outcome.
The target is to learn a function $\hat{\tau}$ to estimate CATE.
Empirically, with cross-validation, we can evaluate the quality of the learned function $\hat{\tau}(\bm{x})$ based on the mean squared error (MSE):
\begin{equation}
\epsilon_{PEHE} = \frac{1}{n}\sum_{i=1}^n(y_i^1-y_i^0-\hat{\tau}(\bm{x}_i))^2,
\end{equation}
which is often referred to as \textit{precision in estimation of heterogeneous effect} (PEHE). 
It is also adopted for evaluating estimated individual treatment effects (ITE)~\cite{hill2011bayesian,louizos2017causal,johansson2016learning,shalit2017estimating}. Note that PEHE is the mean squared error of the estimated ITEs.

\subsection{Traditional Methods without Unobserved Confounders}
\label{subsec:SD}
\begin{figure}[tbh!]
		\centering
	\begin{tikzpicture}

\node[state] (d) at (0,0) {$t$};

\node[state] (x) at (1.5,0.5) {$\mathbf{x}$};

\node[state] (y) at (3,0) {$y$};

% Directed edge
\path (x) edge (d);
\path (x) edge (y);
\path (d) edge[bend left=0] (y);

\end{tikzpicture}
	\caption{A causal graph for the unconfoundedness assumption which is used for learning causal effects.}
	\label{fig:CI_S}
	\vspace{-5pt}
\end{figure}
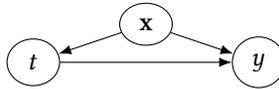

To simplify the problem, it is commonly assumed (or hoped) that all confounders are among the observed features.
In SCMs, this is equivalent to assuming that conditioning on some subset of observed features, denoted by $\bm{s}$, can block all the back-door paths.
\emph{Adjustment} eliminates confounding bias based on the subset of features $\bm{x}$.
We introduce three families of adjustments: regression adjustment, propensity score methods and covariate balancing.
We assume binary treatment $t\in\{0,1\}$ and adopt the language of generalized structural equation introduced in Section~\ref{sec:model}.
The causal graph embedding the assumption for such methods is shown in Fig.~\ref{fig:CI_S}.

\subsubsection{Regression Adjustment}
\label{subsec:RA}
In supervised learning, we fit a function to estimate the probability distribution $P(y|\bm{x})$ where $y$ and $\bm{x}$ denote the label and the features.
As discussed in Section~\ref{sec:model}, to learn causal effects, we are interested in interventional distributions and counterfactuals which cannot be directly estimated from data.
Following the potential outcome framework, we infer the counterfactual outcomes $y_i^{1-t_i}$ based on the features $\bm{x}$ and the treatment $t$.
Roughly speaking, there are two types of regression adjustment.
One is to fit a single function to estimate $P(y|\bm{x},t)$. 
It is enough for inferring ITE because there would be no confounding bias by conditioning on $\bm{x}$, i.e., $P(y|t,\bm{x})=P(y|do(t),\bm{x})$. 
So we can infer the counterfactual outcome as $\hat{y}_i^{1-t_i} = \mathds{E}(y_i|1-t_i,\bm{x}_i)$.
The second is to fit a model for each potential outcome, i.e. $P^1(y|\bm{x}) = P(y|t=1,\bm{x})$ and $P^0(y|\bm{x}) = P(y|t=0,\bm{x})$.
Then we can estimate ATE by:
\begin{equation}
\label{eq:est_ATE}
	\hat{\tau} = \left [{\sum_{i=1}^n (\hat{y}_i^1-\hat{y}_i^0)} \right]/{n},
\end{equation}
where we estimate $\hat{y}_i^{t}$ by the model $\mathds{E}(y|t,\bm{x}_i)$.

\subsubsection{Propensity Score Methods}
\label{subsec:propen}
Propensity score methods can be considered as a special case of matching methods~\cite{morgan2015counterfactuals}.
Matching methods divide instances into strata and treat each stratum as a randomized controlled trial (RCT).
Based on this assumption, ATE is identified and can be estimated by the na\"ive estimator within each stratum.
In matching methods, we assume \textit{perfect stratification}, which means that (1) each group is defined by a set of features $\bm{x}$; (2) instances in a group are indistinguishable except the treatment and the potential outcomes~\cite{morgan2015counterfactuals}.
Formally, perfect stratification means $\mathds{E}[y_i^t|t_i=1,f(\bm{x})] = \mathds{E}[y_i^t|t_i=0,f(\bm{x})], t\in\{0,1\}$. Function $f(\bm{x})$ outputs a continuous value and we stratify instances into groups based on $f(\bm{x})$.
This equation can be interpreted as: given the group affiliation, the expected values of the potential outcomes do not change with the observed treatment.
This is equivalent to the unconfoundedness assumption in each stratum defined by $f(\bm{x})$ whose parameterization can be flexible. 

But when there exists a group which only contains instances with $t=1$ or $t=0$, where we cannot estimate ATE in such a stratum with the na\"ive estimator.
This issue is referred to as a \textit{lack of overlap}. Note that the strong ignorability condition includes $P(t = 1 | \bm{x}) \in (0,1)$, thus ensuring overlap in the large sample limit.
To mitigate the lack of overlap, \textit{matching as weighting} methods are proposed, side-stepping the need for perfect matchings, which are difficult to achieve in practice.
The most widely adopted methods define the function $f(\bm{x})$ as an estimator of the \textit{propensity score} $P(t|\bm{x})$.
Although the features and treatment assignments are fixed given observational data, we assume that observed treatment is assigned by sampling from the true propensity score: $t_i\sim P(t_i|\bm{x}_i)$. That is, the propensity score is the probability of receiving treatment, given the features.
The main advantage of the propensity score, in contrast to perfect matching, is that it is a sufficient dimension reduction \citep{rosenbaum1983central} in the sense that strata defined purely in terms of the propensity score will permit unconfounded causal inference.
Moreover, as a single continuous quantity, it is possible to develop estimators based on weighted averages rather than discrete stratification. 

Of course, in practice the propensity score is not known beforehand, but must be estimated. Fortunately, supervised learning provides many methods for estimating the propensity score by training a classifier to predict whether an instance would be treated, given its features.
It is common to estimate $P(t|\bm{s})$ by logistic regression. Despite its popularity, note that the validity of a linear logistic model can be suspected and nonparametric alternatives are readily available. 

Propensity score methods can be categorized into four classes~\cite{austin2011introduction}:
\textit{propensity score matching} (PSM), \textit{propensity score stratification}, \textit{inverse probability of treatment weighting} (IPTW), and \textit{adjustment based on propensity score}.
Here we focus on the PSM and IPTW as propensity score stratification is an extension of PSM, and adjustment based on propensity score is a combination of regression adjustment and propensity score methods.

\noindent\textbf{Propensity Score Matching (PSM).} PSM matches a treated (controlled) instance to a set of controlled (treated) instances with similar propensity scores.
For example, in \textit{Greedy One-to-one Matching}~\cite{gu1993comparison}, for each instance $i$, we find an instance $j$ with the most similar propensity score to $i$ in the other treatment group.
Once the instances are matched, we can estimate ATE as:
\begin{equation}
\label{eq:PSM_ATE}
\hat{\tau} =
	 \left[{\sum_{i:t_i=1}(y_i-y_j)+\sum_{i:t_i=0}(y_{j}-y_{i})}\right]/{n}.
\end{equation}
Besides the Greedy One-to-one PSM, there are many other PSM methods.
The difference comes what methods we use to match instances.
Readers can check~\cite{austin2011introduction} for various PSM methods.
Stratification on propensity scores is an extension of PSM.
Having propensity score estimated, we can stratify instances based on the predefined thresholds on propensity scores or the number of strata.

Thus, stratum-specific ATE can be calculated by the na\"{\i}ve estimator.
Specifically, ATE is calculated as the weighted average over all strata:
\begin{equation}
\label{eq:PSS}
	\hat{\tau} = {\sum_{j}|U_j|(\frac{1}{|U_j^1|}\sum_{i\in U_j^1}y_i-\frac{1}{|U_j^0|}\sum_{i\in U_j^0}y_i)}/{\sum_{j}|U_j|},
\end{equation}
where $U_j$, $U_j^1$ and $U_j^0$ denote the set of instances, treated instances and controlled instances in the $i$-th stratum, respectively.
A combination of regression adjustment and propensity score stratification can be used to account for the difference between instances in the same stratum~\cite{austin2011introduction,imbens2004nonparametric,lunceford2004stratification}.

\noindent\textbf{Inverse Probability of Treatment Weighting (IPTW).}
IPTW~\cite{hirano2003efficient} is a covariate balancing method.
Intuitively, we can weight instances based on their propensity scores to synthesize a RCT~\cite{austin2011introduction}.
A common way to define the sample weight $w_i$ is by:
\begin{equation}
\label{eq:IPTW_weight}
	w_i = \frac{t_i}{P(t_i|\bm{x}_i)} + \frac{1-t_i}{1-P(t_i|\bm{x}_i)}.
\end{equation}
With Eq.~\ref{eq:IPTW_weight}, we can find that for a treated instance $i$ and a controlled instance $j$, $w_i=\frac{1}{P(t_i|\bm{x}_i)}$ and $w_{j}=\frac{1}{1-P(t_j|\bm{x}_{j})}$.
So the weight refers to the inverse probability of receiving the observed treatment (control).
For example, if we observe 10 instances with $\bm{x}_i=\bm{x}$ and only one of them is treated. Then we estimate the propensity score $P(t=1|\bm{x})$ as 0.1.
To synthesize a RCT, we need to balance the two treatment groups by weighting the treated instance 9 times as the instances under control, which is done by Eq.~\ref{eq:IPTW_weight}. 
Then we can calculate a weighted average of factual outcomes for the treatment and control groups:
\begin{equation}
\label{eq:IPTW_ATE}
	\hat{\tau} = \frac{1}{n^1}\sum_{i:t_i=1} w_i y_i - \frac{1}{n^0}\sum_{i:t_i=0} w_i y_i,
\end{equation}
where $n^1,n^0$ denote the number of instances under treatment and control.
This is based on the idea that weighting the instances with inverse probability makes a synthetic RCT.
Hence, a na\"{\i}ve estimator can be applied to estimate the ATE as in Eq.~\ref{eq:IPTW_ATE}.
Regression adjustment can also be applied to the weighted dataset to reduce the residual of the synthetic RCT~\cite{joffe2004model}.
Instances with propensity score close to 1 or 0 may suffer from an extremely large weight.
In~\cite{hernan2000marginal}, Hernan proposes to stabilize weights to handle this issue in IPTW.

\noindent\textbf{Doubly Robust Estimation (DRE).}
Funk et al.~\cite{funk2011doubly} propose DRE as a combination of a regression adjustment $\mathds{E}[y|t,\bm{x}]$ and another method that estimates the propensity score $\mathds{E}[t|\bm{x}]$.
In fact, only one of the two underlying models needs to be correctly specified to make it an unbiased and consistent estimator of ATE.
In particular, a DRE model estimates individual-level potential outcomes based on these two models as:
\begin{equation}
\label{eq:dre}
\begin{split}
\hat{y}_i^1 = \frac{y_i t_i}{\hat{P}(t_i|\bm{x}_i)}-\frac{\tilde{y}_i^1(t_i-\hat{P}(t_i|\bm{x}_i))}{\hat{P}(t_i|\bm{x}_i)} \quad,\quad
\hat{y}_i^0 = \frac{y_i(1-t_i)}{1-\hat{P}(t_i|\bm{x}_i)}-\frac{\tilde{y}_i^0(t_i-\hat{P}(t_i|\bm{x}_i))}{1-\hat{P}(t_i|\bm{x}_i)}
\end{split}
\end{equation}
where $\tilde{y}_i^{t_i}$ denotes the estimated potential outcomes of the instance $i$ with regression adjustment $\mathds{E}[y|t,\bm{x}]$ and $\hat{P}(t_i|\bm{x}_i)$ is the estimated propensity score for the instance $i$.
Taking a closer look at Eq.~\ref{eq:dre}, we can find that the regression adjustment model is applied to the estimation of counterfactual outcomes as:
$\hat{y}_i^{1-t_i} = \tilde{y}_i^{1-t_i}$,
while more complicated, a mixture of the regression adjustment of propensity score models is developed for the factual outcomes.
Then we can estimate ATE by taking the average over the estimated ITE for all the instances as in Eq.~\ref{eq:est_ATE}.

\noindent\textbf{Targeted Maximum Likelihood Estimator (TMLE)}~\cite{van2006targeted} is a more generalized method than DRE.
ATE can be inferred with TMLE as:
$
\frac{1}{n}\sum_{i=1}^n Q_n^*(1,\bm{x}_i) - Q_n^*(0,\bm{x}_i).
$
To obtain $Q_n^*(t,\bm{x})$, there are three steps: (1) A model $Q_n^0(t,\bm{x})$ is fitted to estimate the factual outcomes with the features and the treatment. (2) A model $g(t=1,\bm{x})$ is fitted for propensity scores $P(t=1|\bm{x})$. (3) Given $Q_n^0$ and $g(t,\bm{x})$, another model is fitted to minimize the mean squared error (MSE) on the factual outcomes.
Assuming $y\in[0,1]$, this is done by learning a new estimator $\bar{Q^{*}_n}(t,\bm{x})$ with parameters $\hat{\epsilon}_0$ and $\hat{\epsilon}_1$:
\begin{equation}
    \bar{Q^{*}_n}(t,\bm{x})\,=\,\text{expit}\left[Q_n^0(t,\bm{x})/(1-Q_n^0(t,\bm{x}))\,+\,\hat{\epsilon_{0}}H_{0}(t,\bm{x})\,+\,\hat{\epsilon_{1}}H_{1}(t,\bm{x})\right],
\end{equation}
where $\text{expit}(a) = \frac{1}{1+\exp(-a)}$, $H_{0}(t,\bm{x})\,=\,-\frac{\mathds{1}(t=0)}{g(t=0|\bm{x})}\;\text{and},\;H_{1}(t,\bm{x})\,=\,\frac{\mathds{1}(t=1)}{g(1|W)}$.

\subsubsection{Covariate Balancing.}
Besides reweighting samples with propensity scores, the \textit{covariate balancing} methods learn sample weights through regression~\cite{kuang2017estimating}.

\noindent\textbf{Entropy Balancing (EB).}
Hainmueller~\cite{hainmueller2012entropy} proposes EB, a preprocessing method for covariate balancing.
The goal is to learn sample weights of the instances under control such that the moments of the two groups are matched.
The weights are learned by minimizing the objective:
\begin{equation}
\begin{split}
        \underset{w_i}{\arg\min} & \; H(\bm{w}) = \sum_{i:t_i=0}d(w_i) \; s.t. \; \sum_{i:t_i=0} w_i c_{ri}(\bm{x}_i) = m_r \; with \; r=1,...,R,
\end{split}
\end{equation}
where $\sum_{i:t_i=0} w_i = 1; \; w_i \ge 0, \forall i \in \{i|t_i=0\}$. $d(\cdot)$ is a distance metric (e.g., KL divergence $d(w_i) = w_i\log(w_i/q_i)$) measuring the distance between the learned weights $\bm{w}$ and base weights $\bm{q}, q_i\ge 0,$ and $ \sum_{i}q_i=1$.
We can use uniform weights $q_i=1/n^0$, where $n^0$ denotes the number of instances under control.
$\sum_{i:t_i=0} w_i c_{ri}(\bm{x}_i)=m_r$ refers to a set of $R$ balance constraints where $c_{ri}(\bm{x}_i)$ is specified as a moment function for the control group and $m_r$ denotes the counterpart of the treatment group.
For example, when $c_{ri}(\bm{x}_i) = (x_{i}^j)^r $, then $\sum_{i:t_i=0} w_i c_{ri}(\bm{x}_i)$ denotes the reweighted $r$-th moment of the feature $x^j$ for the control, and therefore, $m_r$ would contain the $r$-th order moment of a feature $x^j$ from the treatment group.
Compared to other balancing methods, EB allows a large set of constraints such as moments of feature distributions and interactions.
Different from the matching methods, EB keeps weights close to the base weights to prevent information loss.

\noindent\textbf{Approximate Residual Balancing (ARB).} ARB~\cite{athey2018approximate} combines balancing weights with a regularized regression adjustment for learning ATE from high-dimensional data.
ARB consists of three steps. First, the sample weights $\bm{w}$ are learned as:
\begin{equation}
    \underset{\bm{w}}{\arg\min} \; (1-\xi)||\bm{w}||_2^2 + \xi ||\frac{1}{n^1}\sum_{i:t_i=1}\bm{x}_i - \bm{X}_{i:t_i=0}^T\bm{w}||_{\infty}^2 \; s.t. \sum_{i:t_i=0}w_i=1 \; and \; w_i\in[0,(n^0)^{-2/3}],
\end{equation}
where $\bm{X}_{i:t_i=0}$ denotes the feature matrix for the control group.
Then a regularized linear regression adjustment model with parameters $\bm{\beta}$ fitted as:
\begin{equation}
    \underset{\bm{\beta}}{\arg\min} \; \sum_{i:t_i=0}(y_i-\bm{x}^T\bm{\beta})^2 + \lambda((1-\alpha)||\bm{\beta}||_2^2 + \alpha ||\bm{\beta}||_1),
\end{equation}
where $\lambda$ and $\alpha$ are hyperparameters controlling the strength of regularization.
At the end, we can estimate ATE as $\hat{\tau} = \frac{1}{n^1}\sum_{i:t_i=1}y_i - (\frac{1}{n^1}\sum_{i:t_i=1}\bm{x}_i^T\bm{\beta}+\sum_{i:t_i=0}w_i(y_i - \bm{x}_i^T\bm{\beta}))$.
Compared to EB~\cite{hainmueller2012entropy}, ARB handles sparseness of high-dimensional data with lasso and elastic net~\cite{tibshirani1996regression}.

\noindent\textbf{Covariate Balancing Propensity Score (CBPS).} CBPS~\cite{imai2014covariate}, a method robust to misspecification of propensity score model, is proposed to model propensity scores and balance covariate simultaneously.
Assuming the propensity score model $f(\bm{x})$ with parameters $\bm{\beta}$, the efficient Generalized Method of Moments (GMM) estimator is used to learn $\bm{\beta}$:
\begin{equation}
    \underset{\bm{\beta}}{\arg\min} \; \left[\frac{1}{n}\sum_{i}g(t_i,\bm{x}_i)\right]^T\Sigma(\bm{t},\bm{X})^{-1}\left[\frac{1}{n}\sum_{i}g(t_i,\bm{x}_i)\right],
\end{equation}
% where $\bar{g}(\bm{t},\bm{X})$ denotes the sample mean of the moment conditions $\bar{g}(\bm{t},\bm{X}) = \frac{1}{n}\sum_{i}g(t_i,\bm{x}_i)$. 
%
$g(t_i,\bm{x}_i) = \left(\frac{t_if'(\bm{x}_i)}{f(\bm{x}_i)}-\frac{(1-t_i)f'(\bm{x}_i)}{1-f(\bm{x}_i)}, \frac{(t_i-f(\bm{x}_i))f'(\bm{x}_i)}{f(\bm{x}_i)(1-f(\bm{x}_i))}\right)^T$ are the moment conditions.
They are derived from the first order condition of minimizing $-\sum_{i=1}^n t_i\log f(\bm{x}_i)+(1-t_i)\log(1-f(\bm{x}_i))$, the MLE estimator by which we learn $\bm{\beta}$.
Similar to EB~\cite{hainmueller2012entropy}, CBPS combines two methods: covariate balancing and IPTW.
Compared to EB, CBPS models propensity scores.

\subsection{Traditional Methods with Unobserved Confounders}
\label{subsec:SV}
In many real-world problems of learning causal effects, there exist unobserved confounders.
In these cases, the assumption of unconfoundedness is not satisfied.
In the language of SCMs, this means we are not able to block back-door path by conditioning on the features.
Therefore, a family of methods are developed to handle this situation.
The intuition is to utilize alternative information.
Here, we focus on three most popular methods for learning causal effects with unobserved confounders: \textit{instrumental variable methods}, \textit{front-door criterion} (or ``identification by enumeration of mechanism''), and \textit{regression discontinuity design}.

\subsubsection{Instrumental Variable Methods}
\label{subsec:IV}
\textit{Instrumental variables} enable us to learn causal effects with unobserved confounders, which are defined as:
\begin{definition}{\textbf{Instrumental Variable}}
	\textit{Given an observed variable $i$, features $\bm{x}$, the treatment $t$ and the outcome $y$, we say $i$ is a valid instrumental variable (IV) for the causal effect of $t \rightarrow y$ iff $i$ satisfies: (1) $i\not\independent t|\bm{x}$, and (2) $i \independent y|\bm{x}, do(t)$~\cite{angrist1996identification}.}
	\label{def:IV}
\end{definition}
This means a valid IV causally influences the outcome only through affecting the treatment.
In SCMs, the first condition means there is an edge $i\rightarrow t$ or a non-empty set of collider(s) $\bm{x}$ s.t. $i\rightarrow t \leftarrow \bm{x}$ where $\bm{x}$ denotes the features or a subset of features.
The second condition requires that $i\rightarrow t\rightarrow y$ is the only path that starts from $i$ and ends at $y$.
Thus, blocking $t$ makes $i\independent y$.
This implies the \textit{exclusive restriction} that there must not exist direct edge $i\rightarrow y$ or path $i\rightarrow \bm{x}'\rightarrow y$ where $\bm{x}' \subseteq \bm{x}$.
Mathematically, for all $t$ and $i\not=j$, this can also be denoted by $y(do(i),t)=y(do(j),t).$

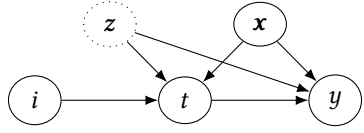
\begin{figure}
	\centering
	\begin{tikzpicture}
	\node[state] (d) at (2,0) {$t$};
	\node[state] (i) at (0,0) {$i$};
	\node[state,dotted] (u) at (1,1) {$\bm{z}$};
	
	\node[state] (x) at (3,1) {$\bm{x}$};

	\node[state] (y) at (4,0) {$y$};
	
	% Directed edge
	\path (x) edge (d);
	\path (x) edge (y);
	\path (d) edge (y);
	\path (i) edge (d);
	\path (u) edge (d);
	\path (u) edge (y);

	% Bidirected edge
	% \path[bidirected] (x) edge[bend left=60] (y);
	\end{tikzpicture}
	\caption{A causal graph of a valid instrumental variable ($i$) when there are unobserved confounders ($\bm{z}$). The binary exogenous variable $i$ stands for whether a customer submits a review. The restaurant type ($x$) is an observed confounder and $\bm{z}$ is a set of unobserved confounders.}
	\vspace{-0.2in}
	\label{fig:IV1}
\end{figure}

In the running example, if we only observe one confounder - the restaurant type ($x$), while the other confounder ($\bm{z}$) remain unobserved.
By assuming that whether a customer submits a review ($i$) is an exogenous random variable, then it is a valid IV (Fig.~\ref{fig:IV1}).
This is because $i$ causally influences $t$ and it can only causally affect $y$ through its influence on $t$.
With a valid IV, we identify the causal effect $t\rightarrow y$ if both the interventional distributions - $P(t|do(i))$ and $P(y|do(i))$ are identifiable.

\noindent\textbf{A Linear SCM for an IV Estimator.}
Here, we show an IV estimator with a linear SCM.
If we also assume that the observed and unobserved confounders $\bm{x}$ and $\bm{u}$ come with a zero mean, we can write down the structural equations for the causal graph in Fig.~\ref{fig:IV1} as:
\begin{equation}
\label{eq:linear_IV}
\begin{split}
	t = \alpha_{i} i + \bm{\alpha}_z^T\bm{z} + \bm{\alpha}_x^T\bm{x} + \alpha_0 + \epsilon_t, \;
	y =  \tau t + \bm{\beta}_z^T\bm{z} + \bm{\beta}_x^T\bm{x} + \beta_0 + \epsilon_y,
\end{split}
\end{equation}
where $\epsilon_t$ and $\epsilon_y$ are Gaussian noise terms with zero mean.
By substituting $t$ in the second equation with the RHS of the first equation in Eq.~\ref{eq:linear_IV}, we get:
\begin{equation}
	y = \tau\alpha_i i + (\tau\alpha_z+\beta_z)^T \bm{z} + (\tau\alpha_x+\beta_x)^T \bm{x} + \gamma_0 + \eta,
\end{equation}
where $\gamma_0=\tau \alpha_0 + \beta_0$, $\eta = \tau \epsilon_d+\epsilon_y$.
Then estimator for the ATE ($\tau$) is obtained:
\begin{equation}
\label{eq:IV_est}
\hat{\tau} =  ({\mathds{E}[y|i]-\mathds{E}[y|i']})/({\mathds{E}[t|i]-\mathds{E}[t|i']}).
\end{equation}
Here, we rely on the following assumptions: linear structural equations, homogeneous treatment effect, valid IV, zero-mean additive noise, and unobserved confounders.
What if some of them do not hold? Can this estimator work under some conditions?
In~\cite{angrist1996identification}, Angrist et al. showed that the ratio estimator (Eq.~\ref{eq:IV_est}) identifies ATE when either the effect of $i$ on $t$ or that of $i$ on $y$ is homogeneous.

\noindent\textbf{An IV Estimator under the potential outcome framework.}
The potential outcome framework formulates the ITE of the IV $i$ on the outcome $y$ as:
\begin{equation}
\label{eq:IV_POF}
\begin{split}
	y_j(i_k=1,t_l(i_k=1)) - y_j(i_k=0,t_l(i_j=0)),
\end{split}
\end{equation}
where $y_j(i_k,t_l)$ and $t_l(i_k)$ are the value of $y$ and $t$ by setting the value of the $k$-th IV to $i_k$. We also assume the IVs are binary.
With the two conditions in Definition~\ref{def:IV}, we know that $i$ affects $y$ via its influence on $t$, so we remove $i_j$ that explicitly influences the value of $y_j$ and reduces Eq.~\ref{eq:IV_POF} to:
\begin{equation}
\begin{split}
		& [y_j^1P(t_j=1|i_j=1)+y_j^0P(t_j=0|i_j=1)] - [y_j^1P(t_j=1|i_j=0)+y_j^0P(t_j=0|i_j=0)] \\
		&= (y_j^1-y_j^0)(P(t_j=1|i_j=1)-P(t_j=1|i_j=0)). \\
\end{split}
\end{equation}
We obtain the ratio estimation (Eq.~\ref{eq:IV_est}) again when we assume homogeneous treatment effect of $i$ on $t$ or that of $i$ on $y$.

\begin{figure}
		\centering
		\begin{tikzpicture}
		\node[state] (d) at (2,0) {$t$};
		\node[state] (i) at (0,0) {$i$};
		\node[state] (s) at (2,1) {$\bm{s}$};
		
		% y node set relative to x.
		% Locations can be:
		% right,left,above,below,
		% above left,below right, etc
		
		\node[state] (y) at (4,0) {$y$};
		
		% Directed edge
		\path (s) edge (i);
		\path (i) edge (d);
		\path (s) edge (y);
		\path (d) edge (y);
		
		\end{tikzpicture}
		\caption{Assuming that we observe all confounders $\bm{x}$ as $\bm{s}$ for the causal effect of the IV, whether a customer writes a review on customer flow $y$, 2SLS can estimate the treatment effect of rating on customer flow ($t\rightarrow y$).}
		\vspace{-0.2in}
	\label{fig:2SLS}
\end{figure}
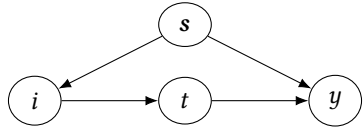

\noindent\textbf{Two-stage Least Square (2SLS).}
As the IV estimator in Eq.~\ref{eq:IV_est} is restrictive, we may have to control a set of features $\bm{x}$ to block the back-door paths between the IV and the outcome so that the IV can be valid.
These cases make it difficult or infeasible to use the estimator in Eq.~\ref{eq:IV_est}.
So we introduce 2SLS~\cite{angrist1995two}.
Fig.~\ref{fig:2SLS} shows an example for such cases where $\bm{x}$ denotes the set of confounders (e.g., whether a coupon can be found on Yelp for the restaurant) for the causal effect of whether a customer makes a review on the customer flow $i\rightarrow y$.
To make $i$ valid, the back-door path $i \leftarrow \bm{x} \rightarrow y$ has to be blocked.
Besides, we may have multiple IVs for each treatment and multiple treatments.
Assuming there is a set of treatments $\bm{t}$ and each treatment $t_{,j}$ has a set of IVs $\bm{i}_{,j}$.
In 2SLS, two regressions are performed to learn the causal effects $(t_{,1}\rightarrow y),...,(t_{,j}\rightarrow y),...$: (1) we fit a function $\hat{t}_{,j} = f_{t_{,j}}(\bm{i}_{,j},\bm{x}_{,j})$ for each treatment variable $t_{,j}$.
	(2) we learn a function $y = g(\hat{\bm{t}},\bm{x})$ where $\hat{\bm{t}}$ signifies the set of treatments.
	Then the coefficient on $D_{,j}$ is a consistent estimate of the ATE of the $j$-th treatment $t_{,j}$ on $y$.
The intuition of 2SLS follows how we find a valid IV.
In the first stage, we estimate how much a certain treatment $t_{,j}$ changes if we modify the relevant IV $i_{,j}$.
In the second stage, we see how the changes in $t_{,j}$ caused by $i_{,j}$ would influence $y$.

{\color{black}For practical studies in big data, Kang et al.~\cite{kang2016instrumental} show identification of causal effects is possible if more than 50\% of the IVs are valid. They also discuss conditions that further allow identification with more than 50\% invalid IVs without knowing which IVs are valid.}

\subsubsection{Front-door Criterion}
\label{subsec:FDC}
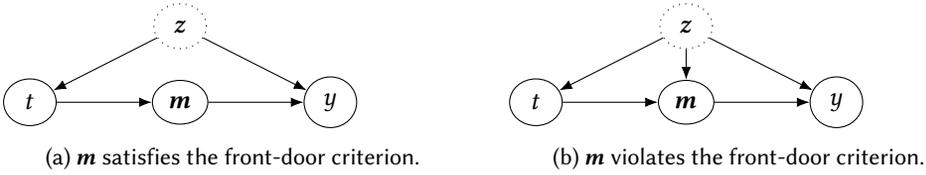
\begin{figure}
	\centering
	\begin{subfigure}[b]{0.44\textwidth}
		\begin{tikzpicture}
		\node[state] (m) at (2,0) {$\bm{m}$};
		\node[state] (d) at (0,0) {$t$};

		\node[state,dotted] (z) at (2,1) {$\bm{z}$};

		\node[state] (y) at (4,0) {$y$};
		
		% Directed edge
		\path (z) edge (d);
		\path (d) edge (m);
		\path (z) edge (y);
		\path (m) edge (y);
		
		\end{tikzpicture}
		\caption{$\bm{m}$ satisfies the front-door criterion.}
		\label{fig:FDC1}
	\end{subfigure}
\hfil
	\begin{subfigure}[b]{0.44\textwidth}
	
		\begin{tikzpicture}
		
		\node[state] (m) at (2,0) {$\bm{m}$};
		
		\node[state] (d) at (0,0) {$t$};
		
		\node[state,dotted] (z) at (2,1) {$\bm{z}$};

		\node[state] (y) at (4,0) {$y$};
		
		% Directed edge
		\path (z) edge (d);
		\path (z) edge (m);
		\path (d) edge (m);
		\path (m) edge (y);
		\path (z) edge (y);

		\end{tikzpicture}
		\caption{$\bm{m}$ violates the front-door criterion.}
		\label{fig:FDC2}
	\end{subfigure}
	\caption{Two causal graphs where $\bm{m}$ satisfies and violates the front-door criterion}
	\vspace{-0.2in}
	\label{fig:FDC}
\end{figure}

The front-door criterion~\cite{pearl1995causal} enables us to learn causal effects $t\rightarrow y$ with unobserved confounders. 
With the front-door criterion we condition on a set of variables $\bm{m}$ which satisfies the following three conditions: (1) $\bm{m}$ blocks all the directed paths from $t$ to $y$,
(2) there are no unblocked back-door paths from $t$ to $\bm{m}$, (3) $t$ blocks all the back-door paths from $\bm{m}$ to $y$.
In other words, we say that the set of variables $\bm{m}$ \textit{mediates} the causal effect of $t$ on $y$.
From the first condition, we decompose $t\rightarrow y$ to a product of $t\rightarrow \bm{m}$ and $\bm{m} \rightarrow y$ as:
$P(y|do(d)) = \int_{\mathcal{M}}P(y|do(\bm{m}))P(\bm{m}|do(d))d\bm{m}.$
The second condition means there is no confounding bias for the causal effect $t \rightarrow \bm{m}$: $P(\bm{m}|do(d)) = P(\bm{m}|d)$.
The third condition infers $P(y|do(\bm{m}))$ by:
\begin{equation}
	P(y|do(\bm{m})) = \int_{\mathcal{T}}P(y|t,\bm{m})P(t)dt.
\end{equation}
Then the interventional distribution corresponding to $t\rightarrow y$ can be identified as:
\begin{equation}
\label{eq:FDC}
	P(y|do(d)) = \int_{\mathcal{M}}P(\bm{m}|d)\sum_{t\in\mathcal{T}}P(y|t,\bm{m})P(t).
\end{equation}
We can estimate the probabilities on the RHS of Eq.~\ref{eq:FDC} from observational data.
For example, we can let the set of variables $\bm{m}$ be the ranking of a restaurant in the search results.
When the ranking is decided by the Yelp rating, ($\bm{z}\independent \bm{x} |t,y$), $\bm{m}$ satisfies the front-door criterion (Fig.~\ref{fig:FDC1}).
However, when the ranking $\bm{m}$ is affected by both the rating $t$ and confounders $\bm{z}$ (e.g. the restaurant category), then $\bm{m}$ is not a valid set of mediators (Fig.~\ref{fig:FDC2}).
Different from the back-door criterion, the front-door criterion enables us to learn causal effects when some confounders are unobserved.

\subsubsection{Regression Discontinuity Design}
\label{subsec:RDD}
Sometimes, treatment assignments may only depend on the value of a special feature, which is the \textit{running variable} $r$.
For example, the treatment is determined by whether its running variable is greater than a cut-off value $r_0$.
The study of the causal effect of Yelp star rating $r$ on the customer flow $y$ is a perfect example for such a case~\cite{Anderson2012}.
Yelp shows the rating of a restaurant rounded to the nearest half star.
For example, restaurant $i$ with average rating 3.26 and restaurant $j$ with 3.24 would be shown with 3.5 and 3.0 stars.
Based on this fact, we can say $r_0=3.25$ is a cut-off which defines the treatment variable.
Then for a restaurant with average rating $R\in[3,3.5]$, we say it receives treatment ($D=1$) when its rounded star rating is greater than its average rating ($R\ge r_0$).
Otherwise, we say a restaurant is under control ($D=0$).
The intuition for Sharp Regression Discontinuity Design (Sharp RDD)~\cite{campbell1969reforms,Anderson2012} is that the restaurants with average rating close to the cutoff $r_0=3.25$ are homogeneous w.r.t. the confounders.
Therefore, what can make a difference in their factual outcomes is the treatment.
In other words, the treatments are randomly assigned to such restaurants, which leads to the identification of the ATE.
In Sharp RDD, we assume that the observed outcome is a function of the running variable as:
\begin{equation}
	\label{eq:Sharp_RDD}
	y_i = f(r_i)+\tau t_i + \epsilon_i = f(r_i)+\tau \mathds{1}(r_i\ge r_0) + \epsilon_{yi},
\end{equation}
where $f(\cdot)$ is a function which is continuous at $r_0$, $\tau$ is the ATE and $\epsilon_{yu}$ denotes the noise term.
The choice of function $f(\cdot)$ can be flexible.
But the risk of misspecification of $f(\cdot)$ exists.
For example, Gelman and Imbens~\cite{gelman2018high} pointed out that high-order polynomials can be misleading in RDD.
In the Yelp study, the fact that customers' decision on which restaurant to go solely relies on the Yelp rating supports this assumption.
For many other real-world problems, however, it is not always the case where we can obtain a perfect cutoff value like the Yelp rating $r_0=3.25$ (stars) and the minimum drinking age $r_0=21$ (years old)~\cite{carpenter2009effect}.
The \textit{Fuzzy RDD} method~\cite{campbell1969reforms,angrist1999using} is developed to handle the cases when cut-offs on the running variable are not strictly implemented.
For example, users may see the real average rating when they look into details of the restaurants and find out that the two restaurants $i$ and $j$ are not that different in terms of rating.
Similar to the propensity score methods, Fuzzy RDD assumes the existence of a stochastic treatment assignment process $P(t|r)$.
But $P(t|r)$ is also assumed to be discontinuous.
The structural equations for Fuzzy RDD is:
\begin{equation}
\label{eq:fuzzy_RDD}
	\begin{split}
	y_if(r_i)+\tau t_i + \epsilon_{yi} = f(r_i) + \pi_2\mathds{1}(r_i>r_0)+\epsilon'_{yi}, \;
	t_i = g(r_i)+\pi_1 \mathds{1}(r_i>r_0) + \epsilon_{ti}
	\end{split}
\end{equation}
where $\tau=\frac{\pi_2}{\pi_1}$ is the ATE we want to estimate, $\epsilon_{yi}$, $\epsilon'_{yi}$ and $\epsilon_{ti}$ are the noise terms.
As $\tau$ is a division between the causal effects $\mathds{1}(r>r_0)\rightarrow t$ and $\mathds{1}(r>r_0)\rightarrow y$, Fuzzy RDD can be considered as an IV method where the discontinuous variable $\mathds{1}(r>r_0)$ plays the role of IV.
A practical guide of RDD can be found in~\cite{cattaneo2017practical}.

{\color{black}
\subsection{Advanced Methods for Learning Causal Effects from Big Data}
\label{subsubsec:adv_methods}
The success of machine learning inspires advanced methods for learning causal effects with big data.
We cover two types of methods: learning causal effects with neural networks and ensembles.

\noindent\textbf{Learning Causal Effects with Neural Networks.}
A straightforward way to learn causal effects with neural networks is to learn representations for features.
To study the causal effect of forming a group on receiving a loan in a microfinance platform, GloVe~\cite{pennington2014glove} and Recurrent Neural Networks (RNN)~\cite{mikolov2010recurrent} embed text features into a low-dimensional space.
Then outcomes are inferred by fitting a function $f(\bm{h},t)$ on factual outcomes $y$.
In~\cite{pham2017deep}, Pham and Shen propose to apply neural networks to estimate the probability distributions such as $\hat{P}(y|t,\bm{x})$ and $\hat{P}(t|\bm{x})$.

Moreover, a series of work learns representations of confounders instead of relying on observed features.
The assumption is that we can learn representations for the confounders, which are considered to be a better approximation of the confounders than the features.
It allows us to go beyond the unconfoundedness assumption.
With specific deep learning models such as the \emph{Balancing Counterfactual Regression}~\cite{johansson2016learning}, the \emph{TARnet}~\cite{shalit2017estimating}, and the \emph{Causal Effect Variational Autoencoder} (CEVAE)~\cite{louizos2017causal}, we can learn representations $\bm{z}_i$ of each instance $i$ based on $(\bm{x}_i,d_i,y_i)$.
Here, we introduce the most recent method, namely the CEVAE, which represents advances along this line.

With the recent advances in variational inference for deep latent variable models, Louizos et al.~\cite{louizos2017causal} propose the CEVAE.
The CEVAE consists of the inference network and the model network.
The inference network is the encoder. Given an instance $(\bm{x}_i, t_i, y_i)$, the encoder learns a multivariate Gaussian distribution $\mathcal{N}(\bm{\mu}_z,\bm{\Sigma}_z) $ from which we can sample its latent representation $\bm{z}_i$.
Then, the model network is the decoder that reconstructs the data from the latent representation.
The two neural networks are shown in Fig.~\ref{fig:CEVAE}.
Those variational distributions ($q(\cdot)$) approximate the corresponding infeasible posterior distributions.
Similar to the VAE~\cite{kingma2013auto} for predictive tasks, the CEVAE is trained through minimizing the KL divergence between the data and its reconstruction.
So the loss function is formulated as:
\begin{equation}
\label{eq:CEVAE_loss}
\mathcal{L} = \sum_{i}	E_{q(\bm{z}_i|\bm{x}_i,t_i,y_i)}[\log P(\bm{x}_i,t_i|\bm{z}_i)+\log P(y_i|t_i,\bm{z}_i)+\log P(\bm{z}_i)-\log q(\bm{z}_i|\bm{x}_i,t_i,y_i)].
\end{equation}
The main difference between the CEVAE and the regular VAE is that, in CEVAE, there is a data point, $(\hat{y}_i^{t},t_i,\hat{\bm{x}}_i)$ reconstructed for each combination of instance and treatment $(i,t)$, which enables the inference of counterfactual outcomes once the neural networks in Fig.~\ref{fig:CEVAE} are trained.
{ \color{black} The comparisons in~\cite{shalit2017estimating,louizos2017causal} with three benchmark datasets (i.e., IHDP, Twins, and Jobs) show that representation learning methods~\cite{shalit2017estimating,louizos2017causal} are the state-of-the-art for learning causal effects.}

\begin{figure}[tbh!]
\tiny
	\begin{subfigure}[b]{0.4\textwidth}
		\centering
	
		\begin{tikzpicture}
	
		\node[state,rectangle] (x) at (0,0) {$P(\bm{x})$};
		
		\node[state,rectangle,dotted] (h11) at (-1.5,1) {hidden};	
		\node[state,rectangle,dotted] (h21) at (-1.5,2) {hidden};
		\node[state,rectangle] (h31) at (-1.5,3) {$q(y|t=1,\bm{x})$};
		\node[state,rectangle] (h51) at (-1.5,4) {$q(\bm{z}|t=1,y,\bm{x})$};
		
		\node[state,rectangle,dotted] (h12) at (1.5,1) {hidden};
		\node[state,rectangle] (h22) at (1.5,2) {$q(t|\bm{x})$};
		\node[state,rectangle] (h32) at (1.5,3) {$q(y|t=0,\bm{x})$};
		\node[state,rectangle] (h52) at (1.5,4) {$q(\bm{z}|t=0,y,\bm{x})$};
		
		% Directed edge
		\path (x) edge (h11) ;
		\path (h11) edge (h21) ;
		\path (h21) edge (h31) ;
        \path (h31) edge (h51) ;
		
		\path (x) edge (h12) ;
		\path (h12) edge (h22) ;
		\path (h22) edge (h32) ;
        \path (h32) edge (h52) ;
		
		\path (h22) edge (h31) ;
		\path (h21) edge (h32) ;

		% Bidirected edge
		\end{tikzpicture}
		\subcaption{The inference network (encoder).}
		\label{fig:CEVAE_infer}
	\end{subfigure}
	\begin{subfigure}[b]{0.4\textwidth}
		\centering
		
				\begin{tikzpicture}
		
		\node[state,rectangle] (x) at (0,0) {$P(\bm{z})$};
		
		\node[state,rectangle,dotted] (h11) at (-2,1) {hidden};	
		\node[state,rectangle,dotted] (h21) at (-2,2) {hidden};
		\node[state,rectangle] (h31) at (-2,3) {$P(\bm{x}|\bm{z})$};
		
		\node[state,rectangle,dotted] (h12) at (0,1) {hidden};
		\node[state,rectangle,dotted] (h22) at (0,2) {hidden};
		\node[state,rectangle] (h32) at (0,3) {$P(y|t=1,\bm{x})$};
		
		\node[state,rectangle,dotted] (h13) at (1.5,1) {hidden};
		\node[state,rectangle,dotted] (h23) at (1.5,2) {hidden};
		\node[state,rectangle] (h33) at (1.5,4) {$P(y|t=0,\bm{x})$};
		
		\node[state,rectangle,dotted] (h14) at (3.5,1) {hidden};
		\node[state,rectangle] (h24) at (3.5,3) {$P(t|\bm{z})$};

		% Directed edge
		\path (x) edge (h11) ;
		\path (h11) edge (h21) ;
		\path (h21) edge (h31) ;
		
		\path (x) edge (h12) ;
		\path (h12) edge (h22) ;
		\path (h22) edge (h32)[bend left=20] ;
		
		\path (x) edge (h13) ;
		\path (h13) edge (h23) ;
		\path (h23) edge (h33) ;
		
		\path (x) edge (h14) ;
		\path (h14) edge (h24) ;
		
		\path (h24) edge[bend left=20] (h32) ;
		\path (h24) edge (h33) ;
		
		\end{tikzpicture}
		\subcaption{The model network (decoder).}
		\label{fig:CEVAE_model}
	\end{subfigure}
\caption{The neural network structures of CEVAE. The parameters, i.e., mean and variance, of each variational distribution $q(\cdot)$, are outputs of the neural network layers below it.}
	\label{fig:CEVAE}
\end{figure}
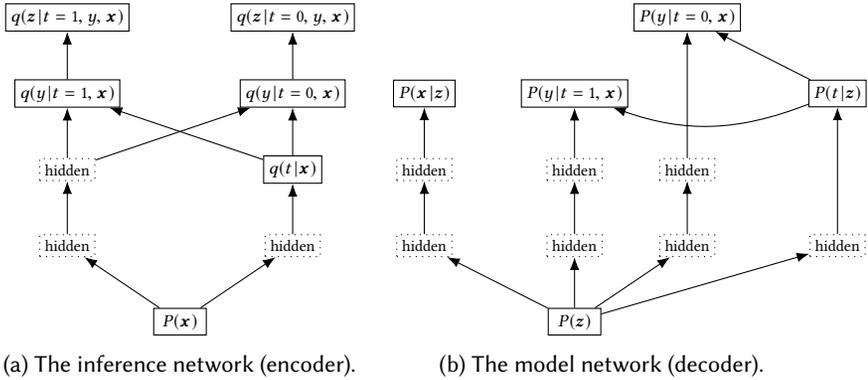

\noindent\textbf{Learning Heterogeneous Causal Effects with Ensembles.}
Ensemble models achieve the state-of-the-art performance in many supervised learning problems.
With ensemble models, we can train a series of weak classifiers on random subsamples of data (i.e., Bootstrapping) and make predictions by aggregating their outputs (i.e., Bagging).
Variants of ensemble models are developed toward learning causal effects.
In~\cite{hill2011bayesian}, Hill proposes to apply Bayesian Additive Trees (BART)~\cite{chipman2010bart} to estimate CATE.
In particular, BART takes the features and the treatment as input and output the distribution of potential outcomes as $f(\bm{x},t) = \mathds{E}[y|t,\bm{x}]$, which returns the sum of the outputs of $Q$ Bayesian regression trees as $f(\bm{x},t) = \sum_{j=1}^Q g_j(\bm{x},t)$.
Thus, we can estimate the CATE for given $\bm{x}$ as $\hat{\tau}(\bm{x}) = f(\bm{x},1)-f(\bm{x},0)$.
Each subtree is defined by the tree structure and a set of $b$ leaf nodes $\left\{\mu_{j1},...,\mu_{jb}\right\}$.
An example of a BART subtree is shown in Fig.~\ref{fig:BART}, where each interior node (rectangle) sends an instance to one of its children.
The $k$-th node of the $j$-th subtree has a parameter $\mu_{jk}$, i.e., the mean outcome of the instances classified to this node.
BART has several advantages~\cite{hill2011bayesian,hahn2017bayesian}: (1) it is good at capturing non-linearity and discontinuity, (2) it needs little hyperparameter tuning and (3) it infers posterior distribution of outcomes, which allows uncertainty quantification.
Hahn et al.~\cite{hahn2017bayesian} propose to handle the problem regularization-induced confounding (RIC) with BART~\cite{hahn2018regularization}.
RIC happens when the potential outcomes heavily depend on the features rather than the treatment.

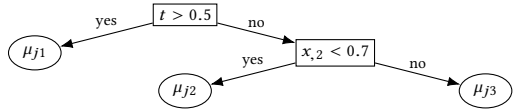
\begin{figure}[t]
\tiny
	\begin{subfigure}[b]{0.4\textwidth}
		\centering
		\begin{tikzpicture}
		
		\node[state] (n2) at (0,0) {$\mu_{j1}$};

		\node[state,rectangle] (n1) at (2,0.5) {$t>0.5$};
		\node[state] (n4) at (2,-0.5) {$\mu_{j2}$};
		\node[state] (n5) at (6,-0.5) {$\mu_{j3}$};
		
		\node[state,rectangle] (n3) at (4,0) {$x_{,2}<0.7$};
	
		% Directed edge
		\path (n1) edge node[above] {yes} (n2) ;
		\path (n1) edge node[above] {no} (n3);
		\path (n3) edge node[above] {yes} (n4);
		\path (n3) edge node[above] {no} (n5);
		\end{tikzpicture}
	\end{subfigure}\\
	\caption{A subtree $g(\bm{x},t)$ in BART}
	\label{fig:BART}
	\vspace{-10pt}
\end{figure}

In~\cite{wager2017estimation}, Wager and Athey propose the Causal Forest which outputs asymptotically normal and consistent estimation of CATE.
Each tree in the causal forest partitions the original covariate space recursively into subspaces such that each subspace is represented by a leaf.
Function $L_j(\bm{x})$ returns which leaf of the $j$-th causal tree in the forest a certain instance belongs to, given its features $\bm{x}$.
Then leaf of the $j$-th tree is considered as a RCT such that the CATE of a given $\bm{x}$ is identified and can be estimated by $\hat{\tau}_j(\bm{x}) = \frac{1}{|U_l^1|}\sum_{i\in U_l^1} Y_i - \frac{1}{|U_l^0|}\sum_{i\in U_l^0} Y_i$,
where $U_l^t=\{i|t_i=t,L_j(\bm{x}_i)=l\}$ refers to the subset of instances that are sent to the $l$-th leaf of the $j$-th subtree whose treatment is $t$.
Then the causal forest simply outputs the average of the CATE estimation from the $J$ trees as $\hat{\tau}(\bm{x}) = \frac{1}{J}\sum_{j}\hat{\tau}_j(\bm{x})$.
Note that there are studies dealing with the case where heterogeneous subpopulations cannot be identified by features such as \textit{principle stratification}~\cite{frangakis2002principal,vanderweele2011principal}.

Sometimes, an instance's treatment or outcome depends on other instances. For example, the customer flow of restaurants in the same area may amount to a constant.
Besides features, treatments, and outcomes, information such as networks, temporal sequential structures can be utilized to capture such dependencies.
Learning causal effects with non-i.i.d data can be done by modeling interference~\cite{rakesh2018linked} and disentangling instances via i.i.d. representations~\cite{guo2019learning,guo2020counterfactual,guo2020ignite}.

While this survey focuses on observational studies, A/B tests play a crucial role in tasks like estimating ATEs, exploring design space and attribute effects to causes for practical decision making~\cite{bakshy2014designing}.
Bakshy et al.~\cite{bakshy2014designing} developed a domain-specific language, PlanOut, for deploying Internet-scale A/B tests.
For A/B tests, big data raises unique challenges.
Taddy et al.~\cite{taddy2016scalable} pointed out the problems that impede using standard tools for inferring ATEs in heavy-tailed distributions (e.g., Internet transaction data): slower learning rate, invalid Gaussian assumption and the failure of nonparametric bootstrap estimators on sampling uncertainty about the mean.
They propose a semi-parametric model for the data generating process of heavy-tailed data to address these issues.
}
\section{Causal Discovery: Learning Causal Relations}
\label{sec:dis}
\begin{figure}
	\centering
	\tiny
	\begin{subfigure}[b]{0.44\textwidth}
		
		\begin{tikzpicture}
		
		\node[state] (d) at (0,0) {$x_{,2}$};
		
		\node[state] (z) at (2,0) {$x_{,1}$};

		\node[state] (y) at (4,0) {$x_{,3}$};
		
		% Directed edge
		\path (z) edge (d);
		\path (z) edge (y);
		
		% Bidirected edge
		\end{tikzpicture}
		\caption{A graph has $x_{,2}\independent x_{,3}|x_{,1}$}
		\label{fig:equi1}
	\end{subfigure}
	\hfil
	\begin{subfigure}[b]{0.44\textwidth}
		\begin{tikzpicture}
\node[state] (d) at (0,0) {$x_{,2}$};
\node[state] (z) at (2,0) {$x_{,1}$};

\node[state] (y) at (4,0) {$x_{,3}$};

% Directed edge
\path (d) edge (z);
\path (z) edge (y);
\end{tikzpicture}
\caption{A graph has $x_{,2}\independent x_{,3}|x_{,1}$}
\label{fig:equi2}
	\end{subfigure}
	\caption{Two exemplary causal graphs that belong to an equivalence class}
	\label{fig:equi}
	\vspace{-10pt}
\end{figure}
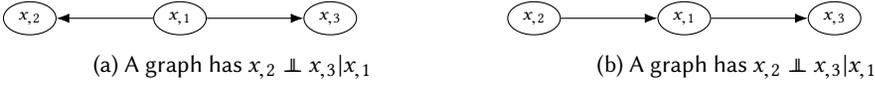

In this section, we start with the problem statement and evaluation metrics. 
{\color{black} Then a review of traditional causal discovery methods followed by methods for causal discovery from big data.}
In learning causal relations (causal discovery), we examine whether a causal relation exists.
\begin{definition}{\textbf{Learning Causal Relations.}}
	\textit{Given $J$ variables, $\{\bm{x}_{,j}\}_{j=1}^J$, we aim to determine whether the $j$-th variable $x_{,j}$ changes if we modify the $j'$-th variable $x_{,j'}$ for all $j\not = j'$.}
\end{definition}
In the running example, learning causal relations enable us to answer the questions such as: \textit{Do features such as location causally affect the customer flow? Is location a confounder for the causal effect of Yelp rating on customer flow?}
To achieve this, we postulate that causality can be detected amongst statistical dependencies~\cite{scholkopf2012causal,peters2017elements}.
An algorithm solving this problem learns a set of causal graphs as candidates~\cite{spirtes2000causation}.
To evaluate the learned causal relations, we often compare each of the learned causal graphs $\hat{G}$ with the ground-truth $G$.
The concept of the \textit{equivalence class} is important for comparing different causal graphs.
\begin{definition}{\textbf{Equivalence Class.}}
	\textit{We say that two causal graphs $G$ and $G'$ belong to the same equivalence class iff each conditional independence that $G$ has is also implied by $G'$ and vise versa.}
\end{definition}
Figs.~\ref{fig:equi1} and~\ref{fig:equi2} show two causal graphs that belong to the same equivalence class. They share the same set of conditional independence $\left\{x_{,2}\independent x_{,3}|x_{,1}\right\}$.

{\color{black}
\noindent\textbf{Evaluation Metrics.}
Here, we briefly introduce the evaluation metrics adopted by the field of learning causal relations.
The metrics can be categorized into two types: (1) the distances between the learned causal graph and the ground truth and (2) the accuracy of discovered causal relations. 
For the first category, a comprehensive study of metrics comparing learned causal graphs (Bayesian networks) can be found in~\cite{de2009comparison}.
In~\cite{chickering2002optimal}, Chickering et al. counted the number of learned causal graphs $G'$ that have the ground truth $G$ as a subgraph.
The structural Hamming distance (SHD) has been widely adopted in~\cite{tsamardinos2006max,peters2016causal}. SHD is defined as the number of edits (adding, removing or reversing an edge) that have to be made to the learned graph $G'$ for it to become the ground truth $G$.
In~\cite{shimizu2011directlingam}, the distance between two graphs are measured by the Frobenius norm of the difference between their adjacency matrices.
The second type of metrics are introduced based on the fact that the discovery of an adjacency relation and an arrowhead can be treated as a binary classification problem. In~\cite{andrews2019learning}, the precision, recall (true positive rate) and false positive rate for both adjacencies (discovered neighbors) and arrowheads (direction of discovered causal relations) are defined as:
\begin{equation}
    \begin{split}
    precision & = \frac{TP}{TP+FP}, \;     recall\;(tpr) = \frac{ TP}{TP+FN}, \; fpr = \frac{FP}{TN+FP}
    \end{split}
\end{equation}
The areas under the precision-recall and fpr-tpr (ROC) curves are widely used~\cite{tsamardinos2006max,bacciu2013efficient}. 
}

\subsection{Traditional Methods for Learning Causal Relations}
\label{subsec:CL_IND}

Following~\cite{malinsky2018causal}, we review three families of algorithms: constraint-based, score-based, and those based on functional causal models. 
The first two rely on statistical tests to discover candidate causal graphs, and the third learns causal relations by estimating coefficients in structural equations.

\noindent\textbf{Constraint-based (CB) Algorithms} learn a set of causal graphs that satisfy the conditional independence embedded in the data.
Statistical tests are utilized to verify if a candidate graph satisfies the independence based on the \textit{faithfulness} assumption~\cite{spirtes2000causation}:
\begin{definition}{\textbf{Faithfulness.}}
	\textit{Conditional independence between a pair of variables, $x_{,j} \independent x_{,j'} | \bm{z}$ for $ x_{,j} \not = x_{,j'}, \bm{z}\subseteq\bm{x}\setminus\left\{x_{,j}, x_{,j'}\right\}$, can be estimated from a dataset $\bm{X}$ iff $\bm{z}$ d-separates $ x_{,j}$ and $x_{,j'}$ in the causal graph $G=(\mathcal{V},\mathcal{E})$ which defines the data-generation process of $\bm{X}$.}
\end{definition}
Faithfulness means the statistical dependence between variables estimated from the data does not violate the independence defined by any causal graph which generates the data~\cite{spirtes2000causation}.
The main challenge is the computational cost as the number of possible causal graphs is super-exponential to the number of variables.
Hence, existing algorithms focus on reducing the number of tests.

\noindent\textit{The Peter-Clark (PC) Algorithm.}
The PC algorithm~\cite{spirtes2000causation} works in a two-step fashion. First, it learns an undirected (\textit{skeleton graph}) from data. 
Then, it detects the directions of the edges to return an equivalent class of causal graphs.
It starts with a fully connected graph and the depth $q=0$.
Then for each ordered pair of connected variables $(x_{,j},x_{,j'})$, it tests if the conditional independence $x_{,j}\independent x_{,j'}|\tilde{\bm{z}}$ is satisfied for each $\tilde{\bm{z}}\subseteq \mathcal{N}(x_{,j})$ or $\tilde{\bm{z}}\subseteq \mathcal{N}(x_{,j'})$ of size $q$, where $\mathcal{N}(\cdot)$ is the set of neighbors.
% %
If the conditional independence holds, it removes the edge $(x_{,j},x_{,j'})$ and saves $\tilde{\bm{z}}$ as the separating set of $(x_{,j},x_{,j'})$. 
% %
Once all such edges are removed, the depth $q$ increases by 1 and this process continues till the number of neighbors for each variable is less than $q$.
In the second step, we decide the directions of edges.
We first determine \emph{v-structures}.
For a triple $(x_{,j},x_{,j'},x_{,j''})$ with no edge between $x_{,j}$ and $x_{,j''}$, we make it a v-structure $x_{,j}\rightarrow x_{,j'} \leftarrow x_{,j''}$ iff $x_{,j'}\not \in \tilde{\bm{z}}$, where $\tilde{\bm{z}}$ denotes saved separating set of $x_{,j}$ and $x_{,j''}$.
{\color{black} Note that no new v-structures would be created as a result of edge orientation.}
Then the remaining undirected edges are oriented following the three rules: (1) we orient $x_{,j} - x_{,j'}$ to $x_{,j} \rightarrow x_{,j'}$ if there exists an edge $x_{,j''}\rightarrow x_{,j'}$ and $x_{,j''}$ and $x_{,j}$ are not neighbors.
(2) we orient $x_{,j} - x_{,j'}$ to $x_{,j} \rightarrow x_{,j'}$ if there is a chain $x_{,j} \rightarrow x_{,j''}\rightarrow x_{,j'}$.
(3) we orient $x_{,j} - x_{,j'}$ to $x_{,j} \rightarrow x_{,j'}$ if there are two chains $x_{,j} - x_{,k}\rightarrow x_{,j'}$ and $x_{,j} - x_{,l}\rightarrow x_{,j'}$.

Other CB algorithms include the IC algorithm~\cite{pearl2009causality} and their variants~\cite{kalisch2007estimating,le2015fast}.
However, most standard statistical tests require Gaussian or multinomial distributions.
To overcome these restrictions, novel conditional independence tests are proposed to cover other families of distributions~\cite{fukumizu2008kernel,zhang2012kernel,sejdinovic2013equivalence,ramsey2014scalable}.
To take unobserved confounders into consideration, algorithms such as FCI (fast causal inference) and its extensions~\cite{spirtes1995causal,colombo2012learning} are proposed to search through an extended space of causal graphs.
Moreover, to go beyond observational data, Kocaoglu et al.~\cite{kocaoglu2017cost} considered the problem of designing a set of interventions with minimum cost to uniquely identify any causal graph from the given skeleton. They show the problem can be solved in polynomial time.

{ \color{black} CB algorithms can also be applied to non-i.i.d. data such as time series.
For example, the FCI algorithm has been adapted for time series~\cite{chu2008search,entner2010causal}.
\emph{Time series models with independent noise} (TiMINo)~\cite{peters2013causal}, a robust algorithm based on non-linear independent tests, can avoid discovering false relations with a misspecified model.
TiMINo takes time series as input and outputs a DAG or remains undecided.}

There are two main drawbacks of this family of algorithms.
First, the faithfulness assumption can be violated. For example, with limited samples, independence tests may even contradict each other.
Second, the causal direction between two variables may remain unknown.

\noindent\textbf{Score-based (SB) Algorithms.}
To relax the faithfulness assumption, SB algorithms replace conditional independence tests with the goodness of fit tests.
SB algorithms learn causal graphs by maximizing the scoring criterion $S(\bm{X},G')$ which returns the score of the causal graph $G'$ given data $\bm{X}$.
Intuitively, low scores should be assigned to the graphs which embed incorrect conditional independence.
For goodness of fit tests, two components need to be specified: the structural equations and the score function.
First, we consider the structural equations.
Structural equations are often assumed to be linear with additive Gaussian noise~\cite{chickering2002optimal}, which introduces parameters $\bm{\theta}$.
Each structural equation describes how a variable is causally influenced by its parent variables and a noise term.
The second component is a score function which maps a candidate causal graph to a scalar based given a certain parameterization of structural equations.
The Bayesian Information Criterion (BIC) score~\cite{schwarz1978estimating}
is the most widely adopted metric $S(\bm{X},G') = \log P(\bm{X}|\hat{\bm{\theta}},G') - \frac{J}{2}\log(n),$
where $\hat{\bm{\theta}}$ is the MLE of the parameters, $J$ denotes the number of variables and $n$ signifies the number of instances.
BIC score prefers causal graphs that can maximize the likelihood of observing the data with regularization on the number of parameters and the sample size.
In~\cite{roos2008bayesian}, a similar score function is proposed based on maximum likelihood estimation with a different regularizer.
Moreover, from the Bayesian perspective, with priors over causal graph structure and parameters, posteriors can be used to define scores.
For example, Bayesian Dirichlet score~\cite{heckerman1995learning} assumes Dirichlet prior on parameters for the multinomial distributions of variables.
With the two components fixed, the score of a certain causal graph for a given dataset is well defined.
Then we focus on searching for the causal graphs which provide the best score for a given dataset.
Searching for the causal graph with maximal score, also known as structural learning is both NP-hard and NP-complete~\cite{chickering1994learning,chickering1996learning}.
It is not computationally feasible to score all possible causal graphs exhaustively.
Therefore, heuristics such as GES~\cite{chickering2002optimal} and its extension, Fast GES (FGES)~\cite{ramsey2017million} are proposed to reach a locally optimal solution.
When it comes to interventional data, Wang et al.~\cite{wang2017permutation} propose algorithms to learn causal relations when a mixture of interventional and observational data is given, which are non-parametric and handle non-Gaussian data well.

\noindent\textit{Greedy Equivalence Search (GES).}
Here we introduce GES as an example of SB algorithms.
In~\cite{chickering2002optimal}, assuming discrete variables, the BDeu criterion is used:
\begin{equation}
\label{eq:scorebased}
    S_{BDeu}(G',\bm{X}) = \log\prod_{j=1}^J 0.001^{(r_j-1)q_j}\prod_{k=1}^{q_j} \frac{\Gamma(10/q_j)}{\Gamma(10/q_j+N_{jk})}\prod_{l=1}^{r_j}\frac{\Gamma(10/(r_i q_i)+N_{jkl})}{\Gamma(10/(r_jq_j))},
\end{equation}
where $r_j$ and $q_j$ signify the number of configurations of variable $x_{,j}$ and parent set $\bm{Pa}_j$ specified by the graph $G'$.
$\Gamma(n)=(n-1)!$ is the Gamma function. $N_{jkl}$ denotes the number of records for which $x_{,j}=k$ and $\bm{Pa}_{j}$ is in the $k$-th configuration and $N_{jk}=\sum_{l}N_{jkl}$.
After initialized with the equivalent class of DAG models with no edges, two-stage greedy search is performed.
First, a greedy search is performed only to insert edges.
The insertion operator $Insert(x_{,j},x_{,j'},\bm{z})$ takes three inputs: $x_{,j}$ and $x_{,j'}$ are non-adjacent nodes in the current graph, $\bm{z}$ denotes any subset of $x_{,j'}$'s neighbors that are not adjacent to $x_{,j}$.
The insertion operator modifies the graph by (1) adding the edge $x_{,j}\rightarrow x_{,j'}$ and (2) directing the previous undirected edge $z-x_{,j'}$ as $z\rightarrow x_{,j'}$. It is worth mentioning that undirected edges can result from the equivalent class of graphs.
As a greedy algorithm, in each iteration, for the current graph, we find the triple $x_{,j},x_{,j'},\bm{z}$ leading to the best score (Eq.~\ref{eq:scorebased}) and perform the insert operator until a local maximum is reached.
Then, the second greedy search is performed initialized with the local optimum of the previous phase, only to delete edges.
The delete operator, $Delete(x_{,j},x_{,j'},\bm{z})$ takes two adjacent nodes $x_{,j}$ and $x_{,j'}$ with edge $x_{,j}-x_{,j'}$ or $x_{,j}\rightarrow x_{,j'}$ and $\bm{z}$ denoting any subset of neighbors of $x_{,j'}$ which are also adjacent to $x_{,j}$.
For each iteration, given the current graph, the triple $x_{,j},x_{,j'},\bm{z}$ with the highest score is selected to update the graph with the delete operator.  
GES terminates when the local maximum is reached in the second phase.

{\color{black} Hybrid algorithms~\cite{wong2002hybrid,tsamardinos2006max} that exploit principled ways to combine CB and SB algorithms have also attracted considerable attention. The MMHC algorithm~\cite{tsamardinos2006max} is proposed toward enough scalability for thousands of variables. First, it learns the skeleton of a causal graph using the Max-Min Parents and Children (MMPC) algorithm~\cite{tsamardinos2003algorithms}, which is similar to the CB algorithms.
Then, it orients the edges with Bayesian scoring hill climbing search, which is similar to SB algorithms.}

\noindent\textbf{Algorithms based on Functional Causal Models (FCMs).}
In FCMs, a variable ${x_{,j}}$ can be written as a function of its directed causes $\bm{Pa}_j$ and some noise term $\epsilon_j$ as $x_{,j}=f(\bm{Pa}_j,\epsilon_j)$.
Different from the two families of methods mentioned above, with FCMs, we are able to distinguish between different DAGs from the same equivalent class.
Here, we adopt Linear Non-Gaussian Acyclic Model (LiNGAM)~\cite{shimizu2006linear} as the FCM to introduce algorithms.
The LiNGAM model can be written as:
$
\label{eq:LiNGAM}
	\bm{x} = \bm{A}\bm{x} + \bm{\epsilon}$,
where $\bm{x}$, $\bm{A}$ and $\bm{\epsilon}$ denote the vector of variables, the adjacency matrix of the causal graph~\cite{shimizu2014lingam}, and the vector of noise, respectively. Columns of both $\bm{x}$ and $\bm{A}$ are sorted according to the \textit{causal order} ($k(j)$) of each variable, respectively.
In the LiNGAM model, the task of learning causal relations turns into estimating a strictly lower triangle matrix $\bm{A}$ which determines a unique causal order $k(j)$ for each variable $x_{,j}$.
For example, if a FCM can be specified by a LiNGAM as:
\begin{equation}
	\begin{bmatrix}
	s \\
	d \\
	y \\
	\end{bmatrix}
	=
	\begin{bmatrix}
	0 & 0 & 0\\
	1.2 & 0 & 0 \\
	0.8 & 1.3 & 0 \\
	\end{bmatrix} 
		\begin{bmatrix}
	s \\
	d \\
	y \\
	\end{bmatrix}
	+
	\begin{bmatrix}
	\epsilon_s \\
	\epsilon_d \\
	\epsilon_y \\
	\end{bmatrix},
\end{equation}
then the causal order of the three variables $s,d,y$ is $1,2$ and $3$, respectively.

\noindent\textit{ICA-LiNGAM.}
Based on independent component analysis (ICA)~\cite{hyvarinen2000independent}, the ICA-LiNGAM algorithm~\cite{shimizu2006linear} is proposed to learn causal relations with the LiNGAM model, with which we estimate the matrix $\bm{A}$. 
First, we can rewrite Eq.~\ref{eq:LiNGAM} as $\bm{x} = \bm{B}\bm{\epsilon}$, where $\bm{B} = (\bm{I}-\bm{A})^{-1}$.
As each dimension of $\bm{\epsilon}$ is assumed to be independent and non-Gaussian, it defines the ICA model for the LiNGAM.
Thus we can apply ICA to estimate $\bm{B}$.
Given data $\bm{X}$ of the variables $\bm{x}$, we use ICA algorithm~\cite{hyvarinen2000independent} to obtain the decomposition $\bm{X}=\bm{B}\bm{S}$.
We can learn $\bm{W}=\bm{B}^{-1}$ by maximizing the objective:
\begin{equation}
\begin{split}
   \sum_{j} J_G(\bm{w}_{j}) \; s.t. \; \mathds{E}[(\bm{w}_k^T\bm{x})(\bm{w}_l^T\bm{x})] = \delta_{kl},
\end{split}
\end{equation}
where $J_G(\bm{w}_{i}) = \{\mathds{E}[G(\bm{w}_i^T\bm{x})]-\mathds{E}[G(\bm{v})]\}^2$, $G$ can be any nonquadratic function (e.g., $G(y)=y^4$). $\bm{v}$ denotes samples from a normal distribution $\mathcal{N}(0,1)$ and $\delta_{kl}$ is the magnitude of dependence between the two variables.
Then an initial estimate of $\bm{A}$, namely $\bm{A}'$, is computed based on $\bm{W}$ as $\bm{A}' = \bm{I} - \tilde{\bm{W}}'$. $\tilde{\bm{W}}'$ is obtained by dividing each row of $\tilde{\bm{W}}$ by the corresponding diagonal element. $\tilde{\bm{W}}$ is calculated by finding the unique permutation of rows of $\bm{W}$ which is nonzero on the diagonal.
Finally, to estimate the causal order $k(j)$ for each $x_{,j}$, permutations are applied to $\bm{A}'$ to obtain an estimate of $\bm{A}$ which is as close to a strictly lower triangle matrix as possible.
A main downfall of ICA-LiNGAM is that ICA algorithms may converge to local optima.
To guarantee the convergence to the global optima in a fixed number of steps, Shimizu et al. propose the DirectLiNGAM algorithm~\cite{shimizu2011directlingam}, which also determines $\bm{A}$ through estimating the causal ordering of variables $k(j)$.

Recently, Additive Noise Models (ANMs) are proposed to relax the linear restriction on the relations between variables and the distribution of noise~\cite{hoyer2009nonlinear,hoyer2012causal}.
ANMs also help reduce the search space of causal graph as data normally does not admit two ANMs with conflicts in directions of causal effects~\cite{hoyer2009nonlinear,peters2017elements}.
One step further, Post-nonlinear Models expand the functional space with non-linear relations between the variables and the noise~\cite{zhang2009identifiability}.

{ \color{black} Those algorithms for FCMs (e.g., ICA-LiNGAM) can also be adapted to handle time series data.}
For example, an auto-regressive LiNGAM is proposed to learn causal relations from time series~\cite{hyvarinen2010estimation}.

{ \color{black} 
\subsection{Learning Causal Relations from Big Data}

The challenges raised by big data for learning causal relations include (1) handling high-dimensional data and (2) dealing with large-scale mixed data.

\noindent\textbf{Causal Discovery from High-dimensional Data.}
There are two questions to answer before applying a causal discovery algorithm to high-dimensional data: (i) Is it theoretically consistent in the high-dimensional setting? (ii) Is it scalable for high-dimensional data?
Traditional CB algorithms (e.g., the PC algorithm~\cite{spirtes2000causation} and its variants~\cite{bacciu2013efficient}) have been shown to be applicable to high-dimensional datasets with thousands of variables in terms of both consistency and scalability~\cite{le2013inferring,gao2015learning}.
Following~\cite{bacciu2013efficient}, in~\cite{colombo2014order}, Colombo and Maathuis point out that outputs of CB methods that use the PC algorithm for obtaining the skeleton depend on the order in which variables are given.
This can be a pronounced issue leading to highly variable results in high-dimensional settings.
Therefore, they propose a new skeleton search algorithm which can be widely used in a series of CB algorithms including the PC, FCI, RFCI and CCD algorithms~\cite{colombo2012learning,spirtes2000causation,richardson1996discovery}.

The case is different for SB algorithms as the traditional work leaves the consistency and scalability in the high dimension setting as open problems~\cite{nandy2018high}.
In~\cite{ramsey2017million}, Ramsey et al. examine two modifications to address the scalability problem of the GES algorithm~\cite{chickering2002optimal}: (1) FGES and (2) FGES with Markov Blanket Search (FGES-MB).
FGES-MB modifies the first stage of GES by only adding edges found by two-hop markov blanket search on each variable~\cite{ramsey2017million}. 
Their experiments show the FGES algorithm with BIC score and FGES-MB can maintain high precision and good recall when the causal graph has one million Gaussian variables with average degree 2.
Nandy et al.~\cite{nandy2018high} find that we can theoretically achieve the consistency in high dimensional settings when adaptive restrictions are added to the GES algorithm~\cite{chickering2002optimal} on the search space.
The proposed variant of GES, ARGES, restricts the choice of variable $z$ in the first stage of the GES algorithm to those form v-structures $(x_{,j},x_{,j'},z)$.
Then theoretically, they show that ARGES is consistent in several sparse high-dimensional settings.
Experimental results show ARGES scales to 400 samples of 2,400 variables and outperforms the baselines (MMHC~\cite{tsamardinos2006max}, PC~\cite{spirtes2000causation} and GES~\cite{chickering2002optimal} etc.) measured by the ROC curve.
When it comes to hybrid algorithms, as we mention in Section 4.1, the MMHC algorithm~\cite{tsamardinos2006max} can scale to thousands of variables.

\noindent\textbf{Learning Causal Relations from Mixed Data.}
Big data often contains both continuous and discrete variables, which is referred to as mixed data.
Mixed data poses challenges for learning causal relations as continuous and discrete variables often require different independence tests (for CB methods) and goodness of fit tests (for SB methods).
Raghu et al.~\cite{raghu2018evaluation} performed an extensive empirical study on the CB and SB methods for learning causal relations with mixed data.
To adapt CB algorithms to mixed data, a straightforward solution is to consider the independence tests that handle mixed data such as a multinomial logistic regression and a conditional Gaussian test (CG).
In the experiments, the authors focused on the small sample high dimension setting, which often fits the case of biomedical data.
Empirical results show that using these independence tests for mixed data can significantly boost the recall of discovered causal relations when there are 100 samples for 100 variables.
In~\cite{sedgewick2016learning}, authors find that modeling the joint distribution of both continuous and discrete variables with mixed graphical models (MGMs)~\cite{lee2015learning} as a preprocessing step can improve CB and SB methods with mixed data. Note that the joint distribution implies a graph structure over these heterogeneous variables.
% %
% 
%

%
For scalable causal discovery on high-dimensional large-scale mixed data, Andrew et al.~\cite{andrews2019learning} derived the degenerate Gaussian (DG) score and proved its consistency. With the four metrics: the precision and recall of arrowheads and those of adjacencies, results show that DG outperformed CG~\cite{raghu2018evaluation}, MGM~\cite{lee2015learning} and copula PC~\cite{cui2016copula} on synthetic datasets with 500 variables and 1,000 samples.

}

{At the end, we point out that most of the existing algorithms share a limitation: they can only be applied to discovering the causal relations of variables whose samples have been observed in the training data.
Developing models that can be generalized to new variables remain an open problem.}

\section{Connections to Machine Learning}
\label{sec:ml}
This section covers connections between learning causality and supervised and semi-supervised learning, domain adaptation, and reinforcement learning.
We explore two aspects: How can causal knowledge improve machine learning? How can machine learning help learning causality?
\subsection{Supervised Learning and Semi-supervised Learning}
\noindent\textbf{Supervised Learning.}
From a data perspective, some problems of learning causality can be reduced to supervised learning or semi-supervised learning problems.

Here, we discuss how supervised learning algorithms can help learning causality.
The problem of learning causal relations can be transformed into as a prediction problem once we label the data with causal relations. 
In particular, suppose we are given labeled training data of the form $(c_1,a_1),...,(c_N,a_N)$ where each $c_j$ is an i.i.d. dataset $c_j={(\bm{X}_1,\bm{y}_1),...,(\bm{X}_{N_j},\bm{y}_{N_j})}$ sampled from a joint distribution $P_j(\bm{x},y)$ and each dataset has an additional label $a_j\in ({\rightarrow, \leftarrow})$ describing whether the dataset is \emph{causal} $\bm{x}\rightarrow y$ or \emph{anti-causal} $y\rightarrow \bm{x}$.
{\color{black} Anti-causal means that the label $y$ is the cause of the features $\bm{x}$.}
The main challenge here is to obtain the label of causal direction.
For some datasets, the causal relations are naturally revealed~\cite{lopez2017discovering}.
In addition, we can leverage the knowledge that a dataset is causal or anti-causal to improve supervised learning models.
Causal regularization~\cite{bahadori2017causal,shen2017image} is proposed to learn more interpretable and generalizable models.
In~\cite{bahadori2017causal}, a causal regularizer guides predictive models towards learning causal relations between features and labels.
It is assumed that, besides the predictive model, there is a classifier $c^i = P(x^i \text{ does not cause } y)$ outputs whether a feature $x^i$ causes the label $y$.
Then the objective function of a predictive model with the causal regularizer is formulated as:
\begin{equation}
    \underset{\bm{w}}{\arg\min}\; \frac{1}{n}\sum_{j=1}^n \mathcal{L}(\bm{x}_j,y_j|\bm{w}) + \lambda\sum_{i=1}^m c^i|w^i|,
\end{equation}
where $\mathcal{L}$ denotes the loss function of the predictive model with parameters $\bm{w}$.
Intuitively, the lower the probability of a feature to be a cause, the more penalty will be added to its corresponding weight, which eventually encourages the model to pay more attention to those features that are more likely to be causes of the label.
In~\cite{kuang2018stable}, the following causal regularizer is proposed to set each feature as the treatment and learn sample weights such that the distribution of the two groups can be balanced w.r.t. to each treatment (feature):
\begin{equation}
    \sum_{j=1}^m ||\frac{\bm{X}_{,-j}^T(\bm{w}\odot\bm{I}_{,j})}{\bm{w}^T\bm{I}_{,j}} - \frac{\bm{X}_{,-j}^T(\bm{w}\odot(\bm{e}-\bm{I}_{,j}))}{\bm{w}^T(\bm{e}-\bm{I}_{,j})}||_2^2,
\end{equation}
where $\bm{w}\in \mathds{R}^{n}$ signifies the sample weights, $\bm{e}$ denotes the $n\times 1$ vector with all elements equal to $1$, $\bm{X}_{,j}$ and $\bm{X}_{,-j}$ are the $j$-th column of the feature matrix and the matrix of remaining features, $\bm{I}_{i,j}$ refers to the treatment status of the $i$-th instance when the $j$-th feature is set as the treatment.
The authors added a constraint to the original loss function of a logistic regression model to ensure the value of this causal regularizer is not greater than a predefined hyperparameter $\gamma\in \mathds{R}^+$.
Doing this can help identify causal features and construct robust predictive model across different domains.

\noindent\textbf{Semi-Supervised Learning (SSL).}
A machine learning problem can be either causal or \textit{anti-causal}~\cite{scholkopf2012causal}.
For example, in hand written digits recognition~\cite{lecun1990handwritten}, which digit to write is first determined, then the digit would be represented as a matrix of pixel values.
Such causal structure has implications for SSL.
In SSL, the target is to improve the quality of estimated $P(y|\bm{x})$ with additional unlabeled instances which can provide information of the marginal distribution $P(\bm{x})$.
We can first consider the cases when semi-supervised learning would fail. For example, when $p(\bm{x})$ is a uniform distribution, observing more unlabeled instances provides no information about $P(y|\bm{x})$.
In contrast, in the case of representation learning, if $\bm{h}$ contains hidden causes of the features $\bm{x}$, and the label $y$ is one of the cause of $\bm{x}$, then predicting $y$ from $\bm{h}$ is likely to be easy~\cite{Goodfellow-et-al-2016}.
Specifically, the true data-generating process implies $\bm{h}$ is a parent of $\bm{x}$, and thus, $p(\bm{h},\bm{x}) = P(\bm{h})P(\bm{x}|\bm{h})$.
So the marginal distribution is $P(\bm{x}) = \mathds{E}_{\bm{h}}[P(\bm{x}|\bm{h})]$.
With $P(y|\bm{x}) = \frac{P(\bm{x}|y)P(y)}{p(\bm{x})}$, we know $P(\bm{x})$ directly affects $P(y|\bm{x})$.
Therefore, the causal structure of $P(\bm{x})$ can help the prediction of $P(y|\bm{x})$, which achieves the target of SSL.
However, the number of causes can be extremely large.
For example, a positive Yelp review can result from good service, delicious food, cheap price, or decent restaurant environment.
Brute force solutions are not feasible as it is often impossible to capture most of the causes.
So, we need to figure out what causes to encode for a certain target $y$.
Criteria such as mean squared error on reconstructed features are used to train autoencoders and generative models, which assumes that a latent variable is salient iff it affects the value of most features.
However, there can be tasks where the label is only associated with few causes.
For example, to predict the customer flow of truck drivers in fast food restaurants near highway, few hidden causes may be useful.
Therefore, the criteria need to be adaptable in accordance with the task.
Generative Adversarial Networks (GAN)~\cite{goodfellow2014generative} are proposed to address this issue for images.
GAN can adapt its criteria s.t. the latent variables (e.g., ears of human in human head images) that only affect the value of few features can also be learned as representations.
Making optimal decisions on which causes we learn representation for is still an open problem.

Janzing and Sch\"{o}lkopf \cite{janzing2015semi} consider a special case of SSL: let $y = f(x)$ and $x,y \in [0,1]$, where $f$ is an unknown anti-causal model. 
Given $n-1$ labeled instances $\{(x_i,y_i)\}_{i=1}^{n-1}$, we seek to infer the label $y_n = f(x_n)$ of an unlabeled instance $x_n$.
In this setting, it is proved that SSL outperforms supervised learning when $P(x)$ and $f$ are dependent and a certain independence between $P(y)$ and $g=f^{-1}$ holds.
The independence $P(y)\independent g$ is assumed and can be defined as $Cov[P(y),\log g'] = 0$, where $g'$ denotes the derivative of $g$ and $\log g', P(x) \in [0,1]$.
Given that, it can be shown that $Cov[P(x),\log f']>0$, which means $P(x)$ contains information of the function $f$ we aim to learn.
In addition, the fact that SSL only works in the anti-causal direction can help learn causal relations~\cite{sgouritsa2015inference}.
As shown above, if the problem is anti-causal, we expect that better knowledge of $P(x)$ helps prediction of $P(y|x)$ as they are dependent.
In contrast, if it is a causal problem, then knowing $P(x)$ barely helps us learn $P(y|x)$.
Therefore, comparing the errors of estimations on $P(y|x)$ and $P(x|y)$ enables us to determine the direction of causality.
In particular, Gaussian process (GP) regression models are trained to estimate $P(x|y)$ as:
\begin{equation}
\begin{split}
        P(x|y,\bm{y}^*)  = \int_{\mathcal{Z},\mathcal{\theta}}P(\bm{z},\bm{\theta},x|\bm{y}^*,y)d\bm{z}d\bm{\theta}
         \approx \int_{\mathcal{Z},\mathcal{\theta}}P(x|y,\bm{y}^*,\bm{z},\bm{\theta})P(\bm{z},\bm{\theta}|\bm{y}^*)d\bm{z}d\bm{\theta},
\end{split}
\end{equation}
where $\bm{z}$ signifies the latent variables and $\bm{y}^* = (y_1,...,y_{n-1})$ are the observed data points. $\mathcal{Z}$ and $\mathcal{\theta}$ are the set of possible values of the latent variables and the model parameters, respectively.
The first factor $P(x|y,\bm{y}^*,\bm{z},\bm{\theta})$ is the supervised GP regression and the second factor $P(\bm{z},\bm{\theta}|\bm{y}^*)$ denotes the posterior distribution over $\bm{z}$ and $\bm{\theta}$ given observed labels $\bm{y}^*$.
Assuming uniform priors for $\bm{z}$ and $\bm{\theta}$, using Bayes's rule $p(\bm{z},\bm{\theta}|\bm{y}^*) = \frac{p(\bm{y}^*|\bm{z},\bm{\theta})p(\bm{z})p(\bm{\theta})}{p(\bm{y}^*)}\propto p(\bm{y}^*|\bm{z},\bm{\theta}) $ which is parameterized by a Gaussian distribution defined by GP-LVM~\cite{titsias2010bayesian}. Thus, we can estimate $P(x|y,\bm{y}^*) = \frac{1}{m}\sum_{i}p(x|y,\bm{y}^*,\bm{z}^i,\bm{\theta}^i)$ with $m$ MCMC samples from $p(\bm{x},\bm{\theta}|\bm{y}^*)$.
In a similar way, we can find that $p(x|y,\bm{y}^*,\bm{z}^i,\bm{\theta}^i)$ is also proportional to a Gaussian distribution defined by GP-LVM.
Thus, we can estimate $P(x|y)$ as mentioned above and $P(y|x)$ by repeating the procedure with $x$ and $y$ swapped.
Finally, log likelihood of the two estimates reveals the causal direction.
However, it is an open problem to scale such approaches (regression for causal discovery) to high-dimensional, noisy or non-i.i.d. data.

There exists a special type of SSL regarding learning causal effects.
Given massive data on $(\bm{x},t)$ but small samples with observed outcomes $y$, Hahn et al.~\cite{hahn2017bayesian} highlighted that information of $P(\bm{x},t)$ can be brought to help the estimation of $P(y | \bm{x}, t)$. 

\subsection{Domain Adaptation}
Domain adaptation~\cite{daume2009frustratingly,blitzer2006domain} studies how to adapt machine learning models trained in some domains to the others.
One application of domain adaptation is to improve prediction accuracy when we have plenty of labeled data from the source domain (e.g., Yelp review) but not for the target domain (e.g., reviews from another website).
Domain adaptation is related to learning causality by \textit{invariant prediction in different domains} \cite{peters2016causal}, {\color{black} assuming that causal relations do not change across domains.}
Given a target variable $y^e$ and $m$ predictor variables $\bm{x}^e=(x_{1}^e,...,x_{m}^e)$ from different domains $e\in \{1,...,E\}$, the goal is to predict the value of $y$.
Invariant prediction assumes that the conditional $P(y|\bm{Pa}_y)$ is consistent for all domains, where $\bm{Pa}_y$ is a set of direct causes of $y$. Formally,
\begin{equation}
P(y^e|\bm{Pa}_y^e) =P(y^f|\bm{Pa}_y^f).
\label{eq:invar}
\end{equation}
The assumption is valid when the distributions are induced by an underlying SCM and the different domains correspond to different interventional distributions where $y$ is not under intervention.
Then we can conclude that (1) invariant prediction is achieved and (2) $\mathcal{Z}^*$ is the set of estimated causes of the target variable $y$. 
In \cite{peters2016causal}, the authors propose a method to estimate $\bm{Pa}_y$.
Assuming that the collection $\mathcal{S}$ consists of all subsets $S$ of features that result in \emph{invariant prediction}, satisfying $P(y^e|x_S^e)=P(y^f|x_S^f)$, the variables appearing in each such set $S$ form the estimated causes of the label $\bm{Pa}_y$.
Finally, a valid subset $S^*$ that achieves the best performance in the source domains is selected as the features for cross domain prediction.
This is because the selected subset is guaranteed to be optimal in terms of domain generalization error.
Due to the independent mechanism assumption~\cite{rojas2015causal,peters2017elements}, the selected subset is also robust against arbitrary changes of marginal distribution of predictors in the target domains. 
Similar results for domain generalization have been obtained through a global balancing approach~\cite{kuang2018stable} and a causal feature selection method~\cite{paul2017feature}. %
They are based on sample re-weighting.
In~\cite{kuang2018stable}, the proposed model, Deep Global Balancing Regression (DGBR), leverages an auto-encoder model to map data into a latent space before reweighting instead of directly reweighting the original samples~\cite{paul2017feature}. 
We summarize the usage of the low-dimensional representations of DGBR in two ways: (1) They are used in the global balancing regularizer, where each variable is successively set as the treatment variable. 
Then we balance all the variables via learning global sample weights. 
(2) We can predict outcomes based on the representations using regularized regression.
The causal regularizer of DGBR is:
\begin{align}
    \begin{gathered}
    \sum_{j=1}^p\bigg\|\frac{\phi(\bm{X}_{,-j})^T(\bm{w}\odot \bm{X}_{,j})}{\bm{w}^T \bm{X}_{,j}}-\frac{\phi(\bm{X}_{,-j})^T(\bm{w}\odot (\bm{e}-\bm{X}_{,j}))}{\bm{w}^T (\bm{e}-\bm{X}_{,j})}\bigg\|_2^2,
\end{gathered}
\end{align}
where $\bm{w}\in \mathds{R}^{n}$ signifies the sample weights, $\bm{e}$ denotes the $n\times 1$ vector with all elements equal to $1$, $\bm{X}_{,j}$ and $\bm{X}_{,-j}$ are the $j$-th column of the feature matrix and the matrix of remaining features.
In addition, the objective function is a weighted loss of logistic regression along with a constraint to limit the value of this causal regularizer not greater than a predefined positive hyperparameter. 
While for prediction under concept drift~\cite{widmer1996learning}, where Eq.~\ref{eq:invar} is violated but the marginal distributions of the predictors remain the same, one may allow apriori causal knowledge to guide the learning process and circumvent the discrepancies between the source and target domains \cite{pearl2011transportability}, a.k.a. \textit{causal transportability}. The study of transportability seeks to identify conditions under which causal knowledge learned from experiments can be reused in different domains with observational data only.
A formal definition of causal transportability can be referred to \cite{pearl2011transportability}. In \cite{bareinboim2012transportability}, the authors further provide a necessary and sufficient condition to decide, given assumptions about differences between the source and target domains, whether transportability is feasible.

\subsection{Reinforcement Learning}
Reinforcement learning (RL)~\cite{sutton1998introduction} is studied for solving sequential decision-making problems.
The key variables in RL are the action $a$, the state $z$, and the reward $y$.
When an agent performs an action, it reaches the next state and receives a reward. 
The \textit{Markov decision process} is often adopted, where the next state $z_{t+1}$ depends on the current state $z_t$ and action $a_t$ and the reward of the next state $y_{t+1}$ is determined by $z_t$, $z_{t+1}$ and $a_t$.
A RL model learns a \textit{policy} $\pi(a_t,z_t)=P(a_t|z_t)$ which determines which action to take given the current state.
The objective is to maximize the sum of the rewards.
In the running example, we can assume that the state $z_t$ represents the location of a restaurant, the action $a_t$ can be moving to a certain place or staying at the same place and the reward is the customer flow $y$.
In each time step, the restaurant owner decides which action to take and then observes the customer flow.
Then the owner will make decisions for the next time step based on whether the customer flow increases or not.

\noindent\textbf{Unobserved Confounders in RL.}
Unobserved confounders raise issues of learning policies for RL models such as multi-armed bandits (MAB)~\cite{bareinboim2015bandits}.
Without knowing the causal model, MAB algorithms can perform as badly as randomly taking an action in each time step.
Specifically, the Causal Thompson Sampling algorithm~\cite{bareinboim2015bandits} is proposed to handle unobserved confounders in MAB problems.
The reward distributions of the arms that are not preferred by the current policy can also be estimated through hypothetical interventions on the action (choice of arm).
By doing this we can avoid confounding bias in estimating the causal effect of choosing an arm on the expected reward.
To connect causality with RL, we view a strategy or a policy in RL as an intervention~\cite{peters2017elements}.

\noindent\textbf{Unbiased Reward Prediction.}
Given trajectories (actions, states and rewards) of an observed policy, we can utilize causal inference methods to predict rewards for another policy, especially for Episodic RL (ERL) problems.
ERL is a subclass of RL where the state is reset to the default value after a finite number of actions.
ERL helps decision-making in many applications such as card games and advertisement placement~\cite{bottou2013counterfactual}.
One popular approach leverages IPTW for predicting reward of ERL models.
In IPTW, a treatment refers to an action and the policy-specific propensity score is defined as the probability to perform the selected action given the observed state.
Particularly, given trajectories produced by running an observed policy $\pi$ $L$ times  $[(a_1(1),z_1(1)),(a_2(1),z_2(1)),...],...,[(a_1(n),z_1(n)),(a_2(n),z_2(n)),...]$, we can estimate the expected sum of rewards of a policy $\tilde{\pi}$ with IPTW as:
\begin{equation}
\hat{\xi}:=\frac{1}{L}\sum_{l=1}^L y(l)\frac{\prod_{k=1}^K\tilde{\pi}(a_k(l)|z_k(l))}{\prod_{k=1}^K\pi(a_k(l)|z_k(l))},
\end{equation}
where $K$ is the number of time steps in each episode.
Improved variants are proposed in~\cite{bottou2013counterfactual}.
\noindent\textbf{RL with Auxiliary Causal Knowledge.}
There is a line of work to improve RL models with causal knowledge as side information~\cite{lattimore2016causal,yabe2018causal}.
Here, we use the Causal Bandit (CB) problem~\cite{lattimore2016causal} as an example.
In this problem, given $J$ binary variables $x_{,1},...,x_{,J}$ and their causal graph but not the causal mechanisms $x_{,j}=f(\bm{Pa}_j,\epsilon_j)$, we aim to find the intervention that is most likely to set a specified variable $x_{,k}$ to $1$.
An intervention is defined as a vector $\bm{a}\in\{*,0,1\}^J$, where $a_j\not=*$ means $x_{,j}$ is set to $a_j$.
For exploration, a CB algorithm learns $\mu(\bm{a})=P(x_{,k}=1|do(\bm{a}))$, the probability that the target $x_{,k}$ is set to $1$ by $\bm{a}$.
For exploitation, a CB algorithm minimizes the regret $R = \mu(\bm{a}^*)-\mathds{E}[\mu(\hat{\bm{a}})]$, where $\bm{a}^*$ is the optimal action and $\hat{\bm{a}}$ denotes the algorithm's selection.
In~\cite{lattimore2016causal}, the parallel bandit (PB) algorithm is proposed to solve this problem with guarantee to outperform non-causal MAB algorithms.
Given total rounds $L$, in first $L/2$ rounds, the PB algorithm collects observational data by doing intervention $\bm{a}=[*,...,*]$. 
Then it analyzes the observational data for each intervention $\bm{a}=do(x_{,j}=x)$ to estimate the reward as $\hat{\mu}(\bm{a}) = \frac{1}{L_{\bm{a}}}\sum_{l=1}^{L/2}\mathds{1}(x_{l,j}=x_{,j}(\bm{a}))$ and probabilities as $\hat{p}_{\bm{a}}=\frac{2L_{\bm{a}}}{L}, \hat{q}_j=\hat{p}_{do(x_{,j}=1)}$, where $L_{\bm{a}} = \sum_{l=1}^{L/2}\mathds{1}(x_{l,j}=x)$ denotes the number of times we observe what $\bm{a}$ could have done in the observational data.
Next, we create the set of rarely observed actions as $\mathcal{A}' = \{\bm{a}|\hat{p}_{\bm{a}}\le \frac{1}{\hat{m}}\}$, where $\hat{m}$ is a threshold defined by the vector of interventional distributions $\bm{q}$.
Then we uniformly sample $\bm{a}\in \mathcal{A}'$ and observe the value of $x_{,k}$.
At the end, we compute $\mathds{E}[x_{,k}]$ of resulting each action as the estimated reward $\hat{\mu}({\bm{a}})$ and select the one with the largest $\hat{\mu}(\bm{a})$.
Other work bridging RL and causality includes causal approaches for transfer learning in RL models~\cite{zhang2017transfer} and data-fusion for reinforcement learners~\cite{forney2017counterfactual}.

At the end, we summarize advantages and disadvantages of causal machine learning. 
The advantages of machine learning with causality include: (1) invariant prediction under environment changes~\cite{peters2016causal,kuang2018stable,paul2017feature}, (2) model generality and interpretability~\cite{bahadori2017causal,shen2017image}, and (3) performance improvement with theoretical guarantee~\cite{lattimore2016causal,yabe2018causal}.
On the other hand, causal machine learning mainly faces the challenges of insufficient amount of data. Causal machine learning algorithms may require ground truth of counterfactuals~\cite{johansson2016learning,shalit2017estimating} or interventional data~\cite{lattimore2016causal,yabe2018causal} for training or evaluation, which can be difficult to collect.

\section{Conclusions and Some Open Problems}
\begin{figure}

\includegraphics[width=1.00\textwidth]{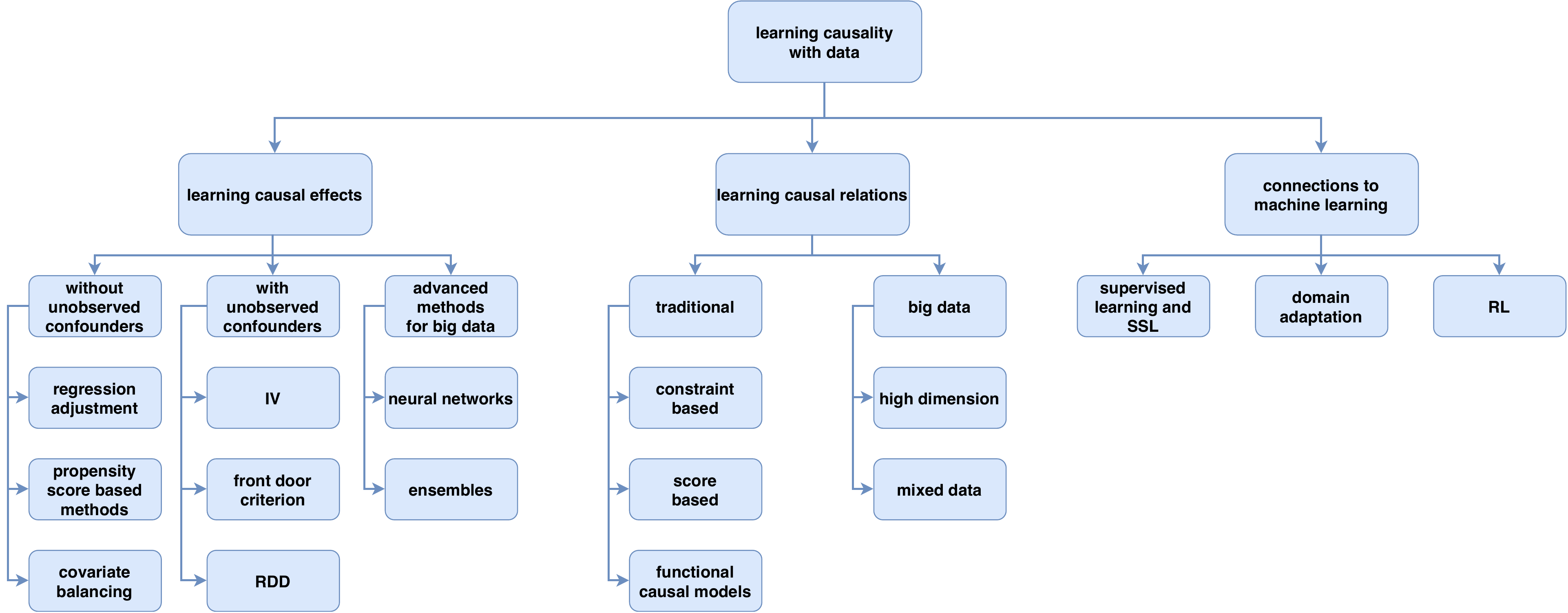}

\caption{Learning causality with data: a summary of the survey}
\vspace{-15pt}
\label{tree:summary}
\end{figure}

Recent attempts have been made to solve the problem of learning causality with more data and less knowledge than traditional studies.
Although existing efforts may not directly address learning causality with big data, they build the foundation of data-driven studies for learning causal effects and relations.
Another highlight of this work is the connections between causality and machine learning.
We aim to demonstrate that it is possible to leverage the connections between them in achieving better solutions for both causal and predictive problems.
Moreover, machine learning models can benefit from exploiting learned causal knowledge in Section~\ref{sec:ml}.

Fig.~\ref{tree:summary} shows a summary of the contents covered in this survey.
SCMs and the potential outcome framework enable us to formulate problems of learning causality with mathematical languages.
Then, we cover the two types of problems: learning causal effect (causal inference) and relations (causal discovery) with data.
The methods for learning causal effects with three types of data are presented: (1) i.i.d. data and (2) non-i.i.d. data where we assume that the unconfoundedness is satisfied, and (3) data with unobserved confounders.
Next, we discuss how to learn causal relationships from two types of data: i.i.d. data and time series data.
Finally, we discuss the connections between learning causality and machine learning.
We discuss how we can connect learning causality to machine learning methods that solve the three families of problems: supervised and semi-supervised learning, domain adaptation and reinforcement learning.
We describe the connections by answering: (1) how learning causality yields better prediction in the machine learning problem and (2) how machine learning techniques can be applied for learning causality?

Existing research demonstrates how to learn causality with data paves the way for more work toward addressing the challenges of big data.
From the data perspective, we present some open problems to achieve the great potential of learning causality with data:
\begin{itemize}
	\item Many problems in learning causality from observational data remain open. Potential problems include handling (1)  anomalies~\cite{akoglu2015graph,ding2019deep}, (2) treatment entanglement~\cite{Toulis2018}, (3) complex treatments~\cite{li2019causal} (e.g., image and text as treatment), (4) temporal observations~\cite{sarkar2019using,marin2017temporal}.

	\item Using causal knowledge to improve machine learning algorithms remains an open area.
	Potential research directions include: (1) causal interpretation of black box deep learning algorithms~\cite{moraffah2020causal}, (2) learning causality-aware models for robustness~\cite{arjovsky2019invariant} and fairness~\cite{kusner2017counterfactual}.
\end{itemize}

\section*{Appendix}
To facilitate development, evaluation and comparison of methods for learning causality, we create the open source data index~\citep{cheng2019practical} (\url{https://github.com/rguo12/awesome-causality-data}) and algorithm index (\url{https://github.com/rguo12/awesome-causality-algorithms}). They are categorized by the problem and the type of data.

\section*{Acknowledgement}

This material is based upon work supported by ARL and the National Science Foundation (NSF) Grant \#1909555 and \#1614576.
\noindent We thank Dr. Judea Pearl for answering Ruocheng Guo's questions on Twitter.
We also thank Dr. Xuan Yin and Dr. Siyu Eric Lu for their invaluable discussion and suggestions.
All remaining errors are Ruocheng Guo's.

% Head 2

% Algorithm

% Enunciations

% Bibliography
\bibliographystyle{ACM-Reference-Format}
\bibliography{sample-bibliography}

\end{document}